\documentclass[10pt,twocolumn,letterpaper]{article}

\usepackage[pagenumbers]{cvpr} 
\usepackage{multirow} 

\usepackage{pifont}

\usepackage{algorithm}
\usepackage{algorithmic}
 \usepackage{amsmath}
 \usepackage{booktabs}
 \usepackage{multirow}
 \usepackage{amsfonts}

\usepackage{pifont}
\newcommand{\cmark}{\ding{51}} 
\definecolor{cvprblue}{rgb}{0.21,0.49,0.74}
\usepackage[pagebackref,breaklinks,colorlinks,allcolors=cvprblue]{hyperref}

\def\paperID{13862} 
\def\confName{CVPR}
\def\confYear{2026}

\title{HeartFormer: Semantic-Aware Dual-Structure Transformers for 3D Four-Chamber Cardiac Point Cloud Reconstruction}

\author{Zhengda Ma$^{\star}$ \qquad
Abhirup Banerjee$^{\star}$\\
$^{\star}$Institute of Biomedical Engineering, Department of Engineering Science, University of Oxford, UK\\
{\tt\small zhengda.ma@eng.ox.ac.uk \qquad abhirup.banerjee@eng.ox.ac.uk}
}

\begin{document}

\maketitle


\begin{abstract}

We present the first geometric deep learning framework based on point cloud representation for 3D four-chamber cardiac reconstruction from cine MRI data. This work addresses a long-standing limitation in conventional cine MRI, which typically provides only 2D slice images of the heart, thereby restricting a comprehensive understanding of cardiac morphology and physiological mechanisms in both healthy and pathological conditions. To overcome this, we propose \textbf{HeartFormer}, a novel point cloud completion network that extends traditional single-class point cloud completion to the multi-class. HeartFormer consists of two key components: a Semantic-Aware Dual-Structure Transformer Network (SA-DSTNet) and a Semantic-Aware Geometry Feature Refinement Transformer Network (SA-GFRTNet). SA-DSTNet generates an initial coarse point cloud with both global geometry features and substructure geometry features. Guided by these semantic-geometry representations, SA-GFRTNet progressively refines the coarse output, effectively leveraging both global and substructure geometric priors to produce high-fidelity and geometrically consistent reconstructions. We further construct \textbf{HeartCompv1}, the first publicly available large-scale dataset with 17,000 high-resolution 3D multi-class cardiac meshes and point-clouds, to establish a general benchmark for this emerging research direction. Extensive cross-domain experiments on HeartCompv1 and UK Biobank demonstrate that HeartFormer achieves robust, accurate, and generalizable performance, consistently surpassing state-of-the-art (SOTA) methods. Code and dataset will be released upon acceptance at: \url{https://github.com/10Darren/HeartFormer}.



\end{abstract}

\section{Introduction}
\label{sec:intro}

\begin{figure}[!h]
\centering
\includegraphics[width=\linewidth]{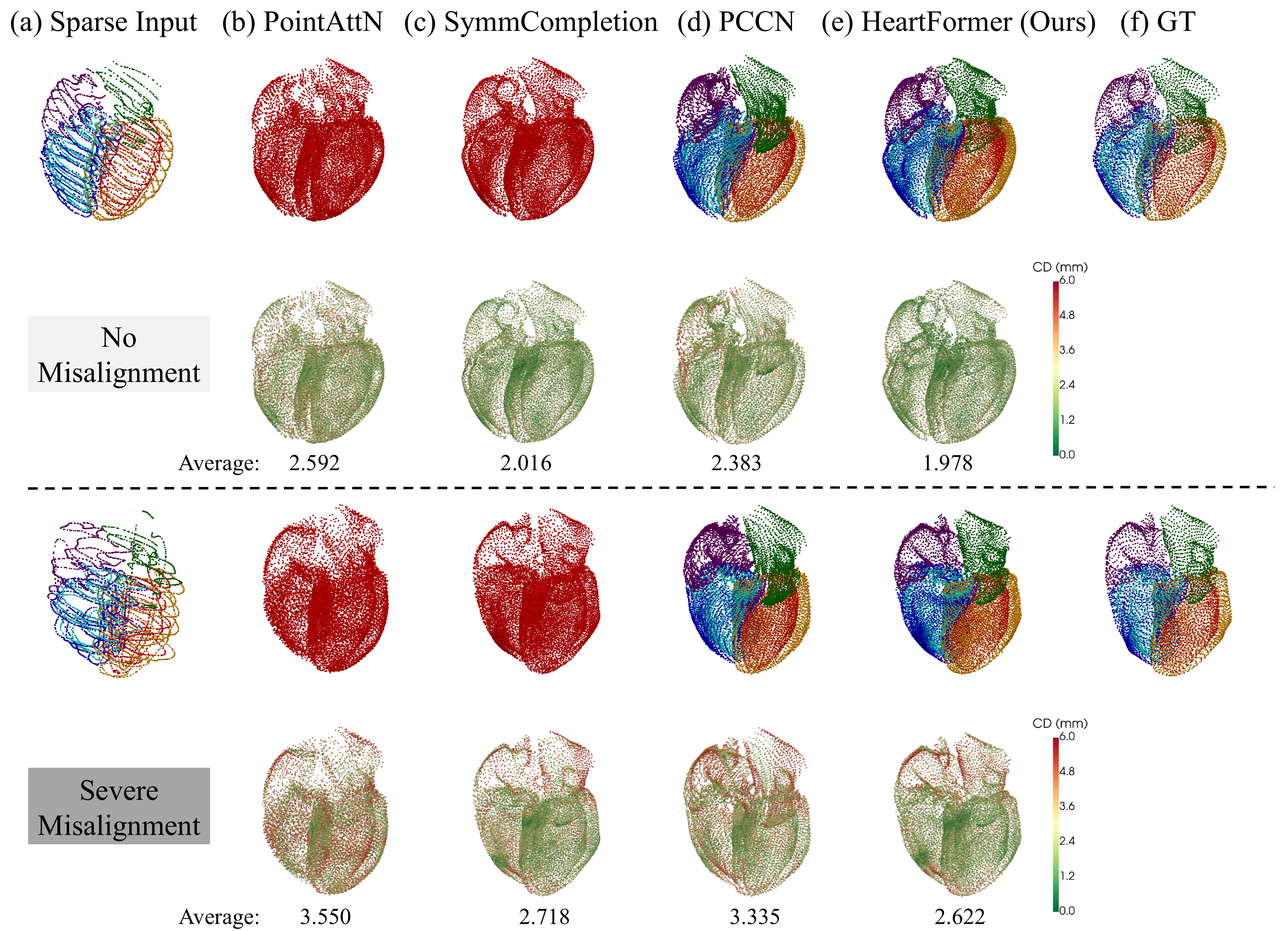} 
\caption{Two sparse point clouds with no and severe misalignment. Qualitative reconstruction results are shown in rows 1 and 3; advanced single-class methods (b,c) tend to generate the entire structure and fail to separate substructures. Rows 2 and 4 visualize the results with point colors indicating the Chamfer distances to the corresponding ground truth (GT) (f). Our proposed HeartFormer (e) achieves significantly better structural fidelity compared to SOTA methods (b-d).}
\label{fig_0}
\vspace{-8mm}
\end{figure}

Cardiac magnetic resonance imaging is widely recognized as the clinical gold standard for evaluating cardiac morphology and function in clinical cardiology \cite{mavrogeni2023cardiovascular, scatteia2023cardiac}. In routine clinical practice, cine MRI acquires a limited set of 2D short- and long-axis slices providing complementary multi-view observations of the heart, yet their inherently 2D nature fundamentally constrains accurate 3D characterization of cardiac anatomy \cite{frangi2002three, beetz2023multi}. High-fidelity 3D representations are essential for quantitative biomarker analysis, pathological visualization, and patient-specific simulations of cardiac mechanics \cite{rodero2023advancing, lashgari2024patient}. Existing methods for 3D cardiac reconstruction predominantly adopt a slice-based segmentation-to-surface paradigm, yet remain constrained by sparse spatial sampling, inter-slice motion, and misalignment \cite{villard2018isachi, banerjee2021completely}. Moreover, recent geometric advances \cite{beetz2022point2mesh,ye2023neural, xiao2024slice2mesh, zhang2025topology} are largely restricted to biventricular model, neglecting atrial anatomy whose thin walls and intricate topology pose additional challenges for accurate reconstruction.

3D point clouds have emerged as a fundamental representation in computer vision and computer graphics, with widespread applications in autonomous driving, augmented reality, and robotics \cite{xiao2023unsupervised}. They offer a flexible and efficient means to describe real-world geometry \cite{huang2024surface}. However, due to the inherent limitations of sensing devices such as LiDAR, depth cameras, and MRI scanners, the captured point clouds are often sparse and incomplete \cite{yuan2018pcn, zhu2025pointsea}. To enhance data quality or enable accurate 3D reconstruction, it is crucial to recover complete and fine-grained structures from partial observations, making point cloud completion a key research problem \cite{zhang2023hyperspherical,chen2023anchorformer}. Consequently, point clouds establish a more direct connection between raw sensor measurements and learning-based geometric reasoning, rendering them particularly advantageous for large-scale scene understanding, 3D reconstruction, and generative modeling \cite{zheng2024point,chen2025parametric}.

Existing point cloud completion networks \cite{yu2301adapointr,wang2024pointattn,rong2024cra,yan2025symmcompletion} primarily focus on single-category synthetic objects (e.g., chairs, cars, tables) or scene-level completion under general 3D scanning setups. In contrast, we introduce the first publicly available multi-class cardiac substructure completion dataset, which aims to simultaneously reconstruct substructures of a single object, i.e. six critical cardiac structures, the left ventricular (LV) endocardium, right ventricular (RV) endocardium, LV epicardium, RV epicardium, left atrium (LA), and right atrium (RA), from incomplete cardiac point clouds. This expands the conventional completion paradigm from geometric restoration to semantic-aware and anatomy-aware multi-class completion.  As illustrated in Fig.~\ref{fig_0}, HeartFormer demonstrates clear advantages over existing approaches in both reconstruction fidelity and anatomical consistency.

Specifically, our contributions as follows:

\begin{enumerate}
\item We introduce the first automated geometric deep learning pipeline for 3D four-chamber heart reconstruction from cine MRI, using 3D point clouds as the intermediate representation.
\item We propose HeartFormer, a unified model that performs multi-class cardiac point cloud completion by leveraging geometric priors and substructure-specific anatomical features, using the proposed Semantic-Aware Dual-Structure Transformer Network (SA-DSTNet). The architecture also integrates a newly designed Semantic-Aware Geometry–Feature Refinement Transformer Network (SA-GFRTNet) that jointly encodes global cardiac geometry and adaptively refines substructure features under semantic-conditioned guidance. Such design enables HeartFormer to handle highly heterogeneous structures within a single model, and the inclusion of semantic-aware loss further ensures structural coherence and reconstruction accuracy.
\item We build HeartCompv1, the first publicly available large-scale dataset comprising 17,000 high-resolution 3D cardiac models (both mesh and point cloud representations).
This dataset establishes the first public resource for systematic evaluation of multi-class medical point cloud completion, promoting fair comparison and large-scale supervised learning in this emerging area.


\item Extensive experiments on the HeartCompv1 and UK Biobank datasets, under cross-domain settings, validate the robustness, accuracy, and generalizability of HeartFormer across both geometric and clinical metrics. The proposed HeartFormer consistently achieves significantly superior performance compared with the SOTA single-class and multi-class point completion methods.

\end{enumerate}



\section{Related work}
\label{sec:Related work}

\begin{figure*}[!h]
\centering
\includegraphics[width=0.9\textwidth]{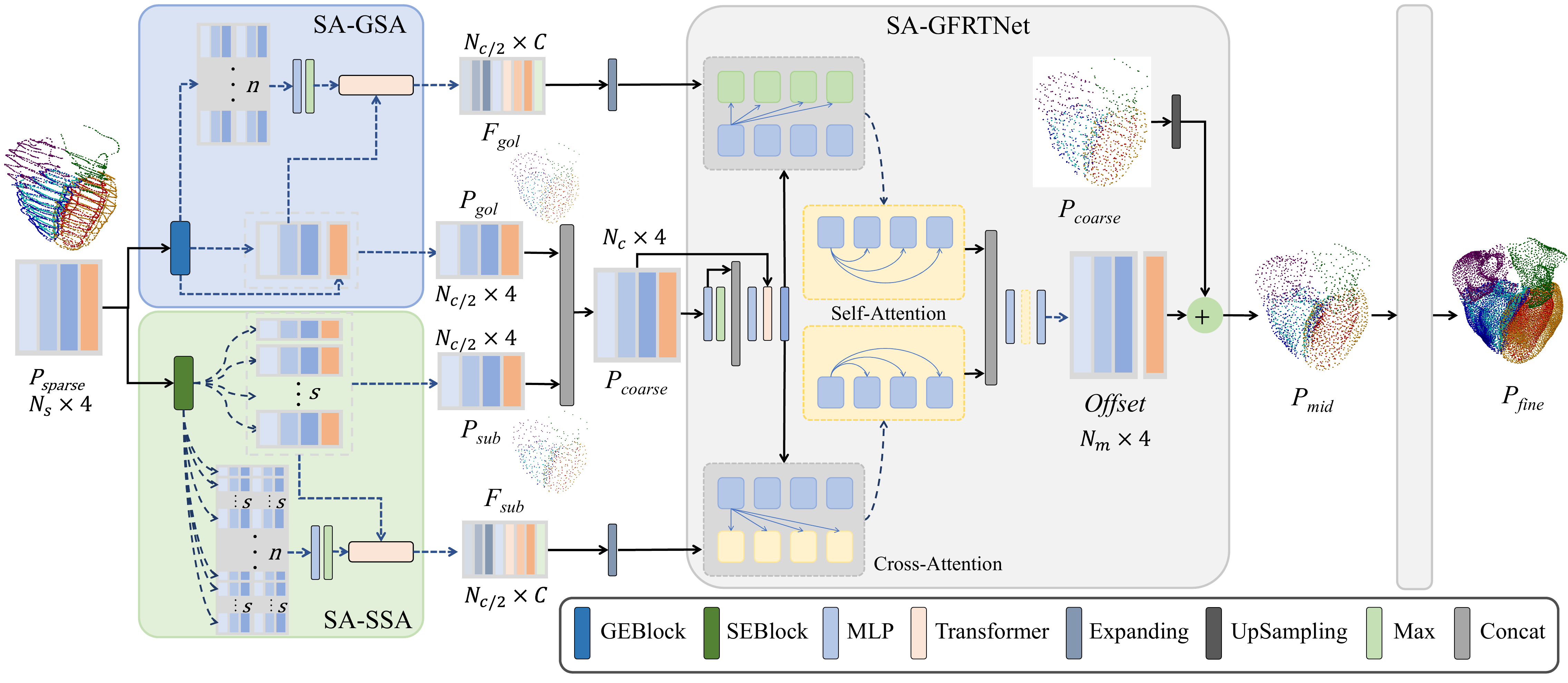} 
\caption{Overview of the proposed HeartFormer architecture. HeartFormer first employs a SA-DSTNet, which integrates a Semantic-Aware Global Structure Aggregation (SA-GSA) module and a Semantic-Aware Substructure Aggregation (SA-SSA) to generate a coarse point cloud, $\mathbf{P}{\text{coarse}}$, by jointly modeling global context and substructure semantics. A subsequent two-stage SA-GFRTNet progressively enhances geometric detail and anatomical consistency, yielding the final high-fidelity reconstruction, $\mathbf{P}_{\text{fine}}$.}
\label{fig_1}
\end{figure*}

\textbf{Point Cloud Completion} is a fundamental problem in 3D vision, which aims to recover dense and high-resolution point clouds from partial observations with low memory overhead \cite{li2025self,wu2024fsc}. With the emergence of deep learning, PCN \cite{yuan2018pcn} was the first to demonstrate the feasibility of directly inferring complete shapes from incomplete point clouds, building upon point-based architectures \cite{qi2017pointnet,qi2017pointnet++,yang2018foldingnet}. Subsequent methods \cite{tchapmi2019topnet,xie2020grnet,huang2020pf,wang2021cascaded} introduced coarse-to-fine generation, hierarchical expansion, and hybrid voxel-point pipelines. Recent advances emphasize fine-grained structural reasoning and efficiency, such as recursive deconvolution (SnowflakeNet \cite{xiang2021snowflakenet}), Transformer-based set translation (PoinTr \cite{yu2021pointr}), and adaptive query mechanisms (FBNet \cite{yan2022fbnet}, AdaPoinTr \cite{yu2301adapointr}, SPD \cite{xiang2022snowflake}). Attention-driven and resolution-aware designs (e.g., AnchorFormer \cite{chen2023anchorformer}, CRA-PCN \cite{rong2024cra}, ODGNet \cite{cai2024orthogonal}, PointAttN \cite{wang2024pointattn}) further enhance contextual aggregation and cross-modal alignment. The latest efforts \cite{yan2025symmcompletion,chen2025parametric} incorporate geometric priors and task-specific objectives, moving toward parametric surface reconstruction and symmetry-aware generation. Beyond solely using partial point clouds, recent approaches exploit combining multi-view depth information \cite{wei2025pcdreamer, zhu2023svdformer, zhu2025pointsea}, image features \cite{aiello2022cross, yu2024geoformer, du2025superpc, liu2025maenet}, textual semantics \cite{yu2024protocomp,li2025genpc}, and template priors \cite{duan2024t} to enhance geometric fidelity, controllability, and structural reliability.

\noindent\textbf{Statistical Shape Model (SSM)} has been widely employed in medical image analysis to characterise anatomical shape variability, playing a pivotal role in 3D reconstruction and patient-specific modelling \cite{heimann2009statistical}. Some approaches \cite{marzola2021statistical, johnson2023application, hoogendoorn2012high} predominantly rely on Principal Component Analysis (PCA) to extract dominant modes of variation from training datasets, enabling compact shape representations in low-dimensional spaces\cite{hoogendoorn2012high}. In recent years, SSM have rapidly evolved toward more automated, high-dimensional, and nonlinear modelling paradigms, with increasing integration of deep learning techniques, thereby significantly enhancing their utility in cardiac shape analysis \cite{banerjee2022automated, beetz2023multi}. Notable examples include variational mesh autoencoders that leverage graph neural networks to learn nonlinear shape embeddings from 3D meshes \cite{beetz2022interpretable}, hybrid approaches that integrate PCA-based SSM with generative adversarial models for congenital heart defect analysis \cite{pandey2025deep}, and end-to-end pipelines achieving sub-2mm reconstruction error in clinically feasible runtimes \cite{govil2023deep}. Furthermore, population-scale studies such as the UK Biobank have demonstrated the broader utility of SSM-derived features as quantitative phenotypes, facilitating genetic association analyses and uncovering novel loci linked to cardiac morphology \cite{burns2024genetic}.


\section{Method}
\label{sec:Method}

This section presents the proposed reconstruction pipeline and our proposed HeartFormer (Sec. 3.1 and 3.2), followed by the synthetic data generation process (Sec. 3.3).

\subsection{3D four-chamber cardiac reconstruction pipeline}

We present the first 3D four-chamber cardiac surface reconstruction pipeline based on point cloud completion, which converts misaligned 2D cine MRI slices into a complete 3D four-chamber cardiac point cloud and mesh. The pipeline takes a cine MRI stack comprising short-axis (SAX) and two- and four-chamber long-axis (2CH, 4CH LAX) views as input. First, a segmentation network generates the corresponding segmentation masks, from which contours are extracted\cite{banerjee2022automated}. These contours, combined with the spatial coordinates, are registered in 3D space and converted into point clouds, resulting in a sparse and misaligned cardiac point cloud. To handle sparse and incomplete points across different substructures, we propose HeartFormer (Sec. 3.2), a multi-class point cloud completion network that simultaneously corrects misalignments and infers dense, anatomically consistent point distributions. The completed point clouds enable high-quality mesh reconstruction through Poisson surface reconstruction followed by Ball Pivoting \cite{bernardini2002ball}. See the supplementary material for an overview of the reconstruction pipeline and implementation details of the segmentation and mesh reconstruction modules.

\subsection{HeartFormer}
\label{sec:HeartFormer}

The overall architecture of our HeartFormer is illustrated in Fig.~\ref{fig_1}. We first employ an SA-DSTNet to generate an initial coarse point cloud, $\mathbf{P}_{\text{coarse}}$, that integrates both global and substructure-level information. Subsequently, a two-stage SA-GFRTNet serves as the core of the coarse-to-fine refinement module, leveraging global context and substructure semantics to progressively refine the point cloud. This process produces a dense, high-fidelity, and anatomically consistent cardiac point distribution, $\mathbf{P}_{\text{fine}}$.

\subsubsection{Semantic-Aware Dual-Structure Transformer Network (SA-DSTNet)}
\label{sec:SA-DSTNet}


We introduce SA-DSTNet to obtain a coarse labeled point cloud that integrates both global context and substructure-level information, while simultaneously extracting corresponding hierarchical features at the global and substructure levels to guide the subsequent refinement process. As illustrated in Fig.~\ref{fig_0}, previous SOTA methods are generally confined to single-object generation, lacking the capability to exploit label information or to explicitly model and analyze substructure relationships. Although multi-class approaches such as PCCN incorporate label information into the input, their output is represented as a $3*K$ (number of classes) dimensional vector, enforcing an equal number of generated points each class. This rigid formulation prevents effective differentiation and representation of individual substructures.
In contrast, our proposed SA-DSTNet effectively overcomes these limitations by jointly leveraging semantic awareness and hierarchical structural reasoning for more precise and adaptive point cloud completion.


As illustrated in Fig.~\ref{fig_1}, the proposed SA-DSTNet comprises two key components: a SA-GSA module and a SA-SSA module. The input sparse point cloud ($\mathbf{P}_{\text{sparse}} \in \mathbb{R}^{N_{\text{s}} \times 4}$) is first processed by SA-GSA to extract semantic-enhanced geometric representations, yielding a set of global key points $\mathbf{P}_{\text{glo}} \in \mathbb{R}^{N_{\text{c}/2} \times 4}$ and their corresponding global features $\mathbf{F}_{\text{glo}} \in \mathbb{R}^{N_{\text{c}/2} \times C}$. In parallel, the SA-SSA processes the input sparse point cloud to capture substructure-level key points $P_{sub} \in \mathbb{R}^{N_{\text{c}/2} \times 4}$ and substructure features $F_{sub} \in \mathbb{R}^{N_{\text{c}/2} \times C}$. $N_{\text{s}}$ and $N_{\text{c}}$ denote the numbers of sparse and coarse points, respectively, $C$ is the feature dimension, and the $4$ encodes 3D coordinates with a semantic label.


\vspace{-10pt}
{\small

\begin{align}
(\mathbf{P}_{\text{glo}}, \mathbf{F}_{\text{1}}) &= \mathcal{F}_{\text{GEBlock}}\big(\mathbf{P}_{\text{sparse}}\big), \\
\mathbf{F}_{\text{glo}} &= \mathcal{T}\!\Big[max_{\text{g}}\!\big(\mathcal{M}_{\text{g}}(\mathbf{F}_{\text{1}}),\, \mathbf{P}_{\text{glo}}\Big], \\
(\mathbf{P}_{\text{sub}}, \mathbf{F}_{\text{2}}) &= \mathcal{F}_{\text{SEBlock}}\big(\mathbf{P}_{\text{sparse}}\big), \\
\mathbf{F}_{\text{sub}} &= \mathcal{T}\!\Big[max_{\text{s}}\!\big(\mathcal{M}_{\text{s}}(\mathbf{F}_{\text{2}}),\, \mathbf{P}_{\text{sub}}\Big],
\end{align}

}
\vspace{-4pt}

\noindent
where \( \mathcal{F}_{\text{GEBlock}} \) and \( \mathcal{F}_{\text{SEBlock}} \) denote the functions of GEBlock and SEBlock, respectively, which extract global and substructure geometric embeddings from the sparse input point cloud \( \mathbf{P}_{\text{sparse}} \). 
The transformer operator \( \mathcal{T}(\cdot) \) further aggregates contextual information to generate the semantic-aware representations \( \mathbf{F}_{\text{glo}} \) and \( \mathbf{F}_{\text{sub}} \). 
The operators \( \max(\cdot) \) and \( \mathcal{M}(\cdot) \) represent the element-wise maximum aggregation and the multi-layer perceptron (MLP), respectively.

Specifically, the GSBlock is composed of a global keypoint sampling layer and a global feature extraction layer, and a jitter layer designed to enhance robustness against misalignment between the input and reference point clouds. Similarly, the SSBlock includes an adaptive substructure keypoint sampling layer and a substructure feature extraction layer. Detailed configurations of both GSBlock and SSBlock can be found in the supplementary.


The global and substructure representations are fused to form the initial coarse point cloud:
\begin{equation}
\mathbf{P}_{\text{coarse}} = [\mathbf{P}_{\text{glo}}, \mathbf{P}_{\text{sub}}],
\end{equation}
where \([\,\cdot\, , \,\cdot\,]\) denotes feature-wise concatenation.

\subsubsection{Semantic-Aware Geometry-Feature Refinement Transformer Network (SA-GFRTNet)}


To further refine both the geometry and the semantic distribution, we introduce an refinement strategy implemented via a two-stage SA-GFRTNet. Learnable offset vectors are applied to the upsampled coarse points, enabling adaptive spatial adjustment and preserving semantic coherence during refinement. In the first stage, the coarse point cloud \( \mathbf{P}_{\text{coarse}} \) is processed by a feature encoding module to obtain its latent representation:
\begin{equation}
    \mathbf{F}_{\text{coarse}} = \Phi_{\text{E}}(\mathbf{P}_{\text{coarse}}),
\end{equation}
where \( \Phi_{\text{E}}(\cdot) \) denotes the feature embedding function. 



In parallel, the global features \( \mathbf{F}_{\text{glo}} \) and substructure features \( \mathbf{F}_{\text{sub}} \) obtained from SA-DSTNet are individually processed by their respective expanding functions. The expanded features, together with the coarse feature \( \mathbf{F}_{\text{coarse}} \), are fed into a Multi-Attention Aggregation (MAA) module that facilitates adaptive information exchange across structural levels.
\begin{equation}
\begin{aligned}
    \mathbf{F}_{\text{glo}}' &= \mathcal{A}_{\text{MAA}}\!\big(\Phi_{\text{g}}(\mathbf{F}_{\text{glo}}),\, \mathbf{F}_{\text{coarse}}\big), \\
    \mathbf{F}_{\text{sub}}' &= \mathcal{A}_{\text{MAA}}\!\big(\Phi_{\text{s}}(\mathbf{F}_{\text{sub}}),\, \mathbf{F}_{\text{coarse}}\big), \\
    \mathbf{F}_{\text{coarse}}' &= [\,\mathbf{F}_{\text{glo}}',\, \mathbf{F}_{\text{sub}}'\,],
\end{aligned}
\end{equation}
where \( \Phi_{\text{g}}(\cdot) \) and \( \Phi_{\text{s}}(\cdot) \) denote the global and substructure expanding functions, respectively, and \( \mathcal{A}_{\text{MAA}}(\cdot) \) represents the MAA operator. 


Given the enhanced feature representation \( \mathbf{F}_{\text{coarse}}' \), an offset generation module predicts fine-grained geometric adjustments, which are then applied to the upsampled coarse points to produce the intermediate point cloud \( \mathbf{P}_{\text{mid}} \):
\begin{equation}
    \mathbf{P}_{\text{mid}} = \mathcal{F}_{\text{offset}}\!\big(\mathbf{F}_{\text{coarse}}'\big)
    + \text{Up}\!\big(\mathbf{P}_{\text{coarse}}\big),
\end{equation}
where \( \mathcal{F}_{\text{offset}}(\cdot) \) denotes the offset generation function and \(\text{Up}(\cdot)\) represents the upsampling operation. 


Subsequently, the intermediate point cloud \( \mathbf{P}_{\text{mid}} \) is fed into the second stage of the SA-GFRTNet, which performs another round of feature aggregation and offset refinement. This hierarchical refinement produces the final output \( \mathbf{P}_{\text{fine}} \):
\begin{equation}
    \mathbf{P}_{\text{fine}} =  \mathcal{A}_{\text{GFRTNet}}(\mathbf{P}_{\text{mid}})
    + \text{Up}\!\big(\mathbf{P}_{\text{mid}}\big),
\end{equation}
where \( \mathcal{A}_{\text{GFRT}}(\cdot) \) denotes the feature aggregation operator within SA-GFRTNet.

\subsubsection{Loss Function}
\label{sec:loss}

For semantics-guided supervision, we employ a Semantic-Aware Chamfer Distance (SA-CD) loss that explicitly integrates anatomical semantics into geometric alignment. 
Unlike the standard Chamfer Distance, which treats all points equally, SA-CD computes the distance independently for each anatomical substructure, ensuring balanced optimization across components of varying sizes and point densities. 
By introducing semantic partitioning, this loss effectively preserves structural boundaries and mitigates cross-class interference during refinement. 
The SA-CD loss is defined as:

\vspace{-10mm}
\begin{equation}
\begin{aligned}
\mathcal{L}_{\text{SA}}(\mathbf{P}, \mathbf{G})
&= \frac{1}{2K} \sum_{k=1}^{K} \Big(
\mathbb{E}_{x \in \mathbf{P}^{(k)}} \min_{y \in \mathbf{G}^{(k)}} \|x - y\|_2 \\
&\quad + \mathbb{E}_{y \in \mathbf{G}^{(k)}} \min_{x \in \mathbf{P}^{(k)}} \|x - y\|_2
\Big),
\end{aligned}
\end{equation}
where $\mathbf{P}^{(k)}, \mathbf{G}^{(k)} \in \mathbf{R}^{N \times 3}$ are the subsets of predicted and GT points corresponding to class $k$. $\mathbb{E}$ denotes the expectation operator. To enforce a coarse-to-fine reconstruction paradigm, SA-CD is applied at all prediction stages:


\vspace{-2mm}
\begin{equation}
\mathcal{L} = \sum_{s \in \{\text{coarse},\, \text{mid},\, \text{fine}\}} 
\mathcal{L}_{\text{SA}}(\mathbf{P}_{s}, \mathbf{G}).
\end{equation}
This hierarchical supervision jointly promotes global structural consistency and locally smooth surface reconstruction across all cardiac substructures.

\subsection{Synthetic dataset generation}

Due to the limited availability of large-scale real 3D whole-heart anatomical datasets, we construct and publicly available  \textbf{HeartCompv1} dataset ($N=17,000$), based on the High-Resolution Atlas and Statistical Model \cite{hoogendoorn2012high}. The dataset comprehensively spans the entire cardiac cycle, capturing both inter-subject anatomical variability and dynamic cardiac motion. Our synthetic data generation pipeline produces sparse, misaligned input point clouds alongside their corresponding dense ground-truth point clouds. Specifically, we sample full heart meshes from the statistical shape model by randomly varying the 50 principal modes of variation. To simulate annotation variability caused by manual selection of the base-apex line of the left ventricle, small random perturbations are introduced to the endpoints, namely the mitral valve center and the left ventricular apex. This augmentation improves the robustness of our method to potential annotation errors.

Next, we define slicing planes for each heart mesh in accordance with standard clinical cine MRI protocols \cite{taylor2005cardiovascular, walsh2007assessment, margeta2014recognizing}. Assuming slice misalignment can be modeled by rigid transformations, we introduce simulated misalignment artifacts caused by respiration or patient motion into each SAX and LAX slice. Following \citet{beetz2023multi}, misalignments are divided into five levels: none, mild, medium, strong, and severe. For each level, independent Gaussian perturbations with increasing standard deviations are applied to reflect varying degrees of misalignment. Random rigid displacements are then applied to each SAX slice and the LAX slices. The resulting 2D contours are converted to 3D point clouds to emulate sparse, misaligned contours found in real cine MRI data. Finally, we extract dense point clouds from the vertices of the deformed heart meshes that serve as ground truth for training. The detailed synthetic data generation protocol is provided in the Supplementary.

\begin{figure*}[!h]
\centering
\includegraphics[width=\textwidth]{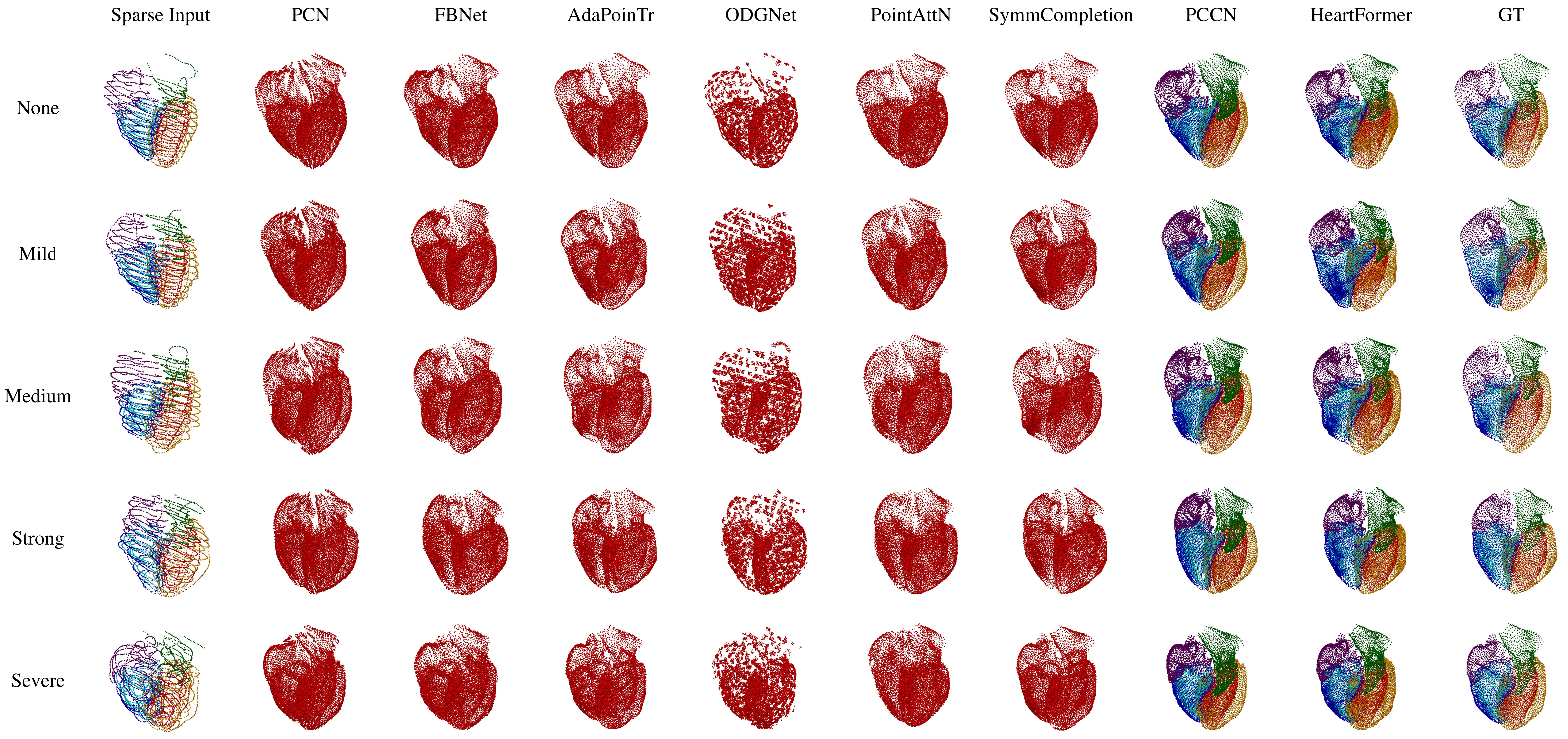} 
\caption{\textbf{Comparison of reconstruction results under five different levels of misalignment in the HeartCompv1 dataset.} Sparse Input denotes the partial point clouds with varying degrees of misalignment. PCN, FBNet, AdaPoinTr, ODGNet, PointAttN, and SymmCompletion are the single-class point cloud completion models, while PCCN and the proposed HeartFormer are multi-class completion models. GT indicates the ground-truth.}
\label{fig_2}
\end{figure*}

\section{Experiments and Results}
\label{sec:Experiments}

\subsection{Experimental Setup}


\noindent\textbf{Implementation Details.} We fix the number of input points to 7,500, and the number of output and ground-truth points to 16,384. All models are trained under the same settings. Training is conducted on four NVIDIA Tesla V100 GPUs using the Adam optimizer with a batch size of 8 and an initial learning rate of 0.0001. Detailed implementation descriptions are provided in the Supplementary Material.

\noindent\textbf{Dataset.} Our primary dataset, HeartCompv1, contains detailed 3D cardiac models (both mesh and point cloud representations) paired with corresponding sparse point clouds for training and evaluation. The training set consists of 10,000 samples with varying degrees of misalignment to improve model robustness. The validation set and six test sets, including five with different levels of misalignment and one with mixed misalignment, each contain 1,000 samples and represent different levels of reconstruction difficulty. In addition, we evaluate the model on a subset of the UK Biobank dataset to assess its generalization in real-world scenarios. Specifically, 116 clinical cases covering both healthy and diseased subjects as well as male and female participants are selected to evaluate model performance under realistic conditions.

\noindent\textbf{Evaluation metrics.} To assess reconstruction accuracy, we employ Chamfer Distance (CD), Hausdorff Distance (HD), and Surface-to-Surface Distance (SSD) as quantitative geometric metrics. In addition, we compute several clinically relevant indices, including standard volumetric measurements at end-diastole: left and right ventricular volumes (LVV, RVV) and left and right atrial volumes (LAV, RAV).

\begin{table}[]
\caption{\textbf{Quantitative results on the mixed-misalignment setting of the HeartCompv1 dataset.} HeartFormer achieves the best overall performance with fewer parameters and lower computational cost. \textbf{Bold} numbers indicate the best results. CD, HD, and SSD are reported in mm; model complexity is measured in Params (M) and FLOPs (G).}
\resizebox{\linewidth}{!}{
\begin{tabular}{l|ccc|cc}
\midrule
Methods        & CD $\downarrow$                  & HD  $\downarrow$          & SSD $\downarrow$           & Params & FLOPs   \\ \midrule
FoldingNet     & 2.399                   & 16.709        & 2.311          & 2.407  & 257.299 \\
PCN            & 2.061                   & 10.365        & 2.242          & 6.864  & 153.524 \\
SnowflakeNet   & 3.357                   & 15.587        & 3.099          & 19.318 & 82.538  \\
PoinTr         & 2.795                   & 20.206        & 1.776          & 31.276 & 85.109  \\
FBNet          & 1.926                    & 10.43         & 2.079          & 4.965  & 221.114 \\
AdaPoinTr      & 1.721                    & 8.449         & 1.765          & 32.493 & 121.042 \\
AnchorFormer   & 5.797                  & 22.712        & 2.524          & 30.464 & 58.481  \\
CRA-PCN        & 2.810                   & 13.522        & 2.432          & 7.468  & 94.413  \\
ODGNet         & 2.247                   & 16.048        & 2.097          & 11.363 & 323.017 \\
PointAttN      & 2.019                   & 10.883        & 2.095          & 22.334 & 157.488 \\
SymmCompletion & 1.657                    & 7.259         & 1.686          & 6.641  & 90.389  \\
PCCN           & 1.949                   & 8.811         & 2.013          & 16.182 & 64.739  \\
\emph{HeartFormer(Ours)}      & \textbf{1.596}  & \textbf{6.75} & \textbf{1.661} & 5.688  & 87.036  \\ \midrule
\end{tabular}
}
\label{tab1_1}
\end{table}

\begin{table*}[]
\centering
\caption{\textbf{Quantitative comparison of reconstruction results for five different levels of misalignment in the HeartCompv1 dataset.}}
\resizebox{0.9\linewidth}{!}{
\begin{tabular}{l|ccccc|ccccc}
\midrule
\multirow{2}{*}{Methods} &                &                & CD $\downarrow$        &                &                &                &                & HD $\downarrow$            &                &                \\
                         & None           & Mild           & Medium         & Strong         & Severe         & None           & Mild           & Medium         & Strong         & Severe         \\\midrule
FoldingNet~\cite{yang2017foldingnet}        & 2.174          & 2.233          & 2.350          & 2.489          & 2.674          & 16.323         & 16.457         & 16.573         & 16.66          & 16.996         \\
PCN~\cite{yuan2018pcn}               & 1.768          & 1.834          & 1.994          & 2.204          & 2.485          & 8.63           & 8.902          & 10.006         & 11.32          & 13.021         \\
SnowflakeNet~\cite{xiang2022snowflake}      & 2.649          & 2.822          & 3.235          & 3.719          & 4.262          & 12.816         & 13.121         & 13.894         & 16.448         & 20.909         \\
PoinTr~\cite{yu2021pointr}            & 2.016          & 2.388          & 2.784          & 3.162          & 3.565          & 18.916         & 19.234         & 19.76          & 20.688         & 23.02          \\
FBNet~\cite{yan2022fbnet}             & 1.453          & 1.710          & 1.939          & 2.147          & 2.369          & 7.738          & 8.446          & 10.141         & 11.788         & 13.722         \\
AdaPoinTr~\cite{yu2301adapointr}         & 1.288          & 1.484          & 1.705          & 1.931          & 2.182          & 7.038          & 7.199          & 8.134          & 9.097          & 10.514         \\
AnchorFormer~\cite{chen2023anchorformer}      & 4.964          & 5.272          & 5.700          & 6.203          & 6.787          & 21.624         & 22.026         & 22.701         & 23.499         & 24.711         \\
CRA-PCN~\cite{rong2024cra}           & 2.158          & 2.384          & 2.705          & 3.108          & 3.600          & 10.596         & 10.992         & 11.827         & 14.399         & 18.979         \\
ODGNet~\cite{cai2024orthogonal}            & 1.743          & 1.942          & 2.186          & 2.488          & 2.839          & 14.523         & 14.901         & 15.657         & 16.525         & 18.759         \\
PointAttN~\cite{wang2024pointattn}         & 1.653          & 1.796          & 1.986          & 2.196          & 2.436          & 8.986          & 9.423          & 10.474         & 11.912         & 13.478         \\
SymmCompletion~\cite{yan2025symmcompletion}    & 1.268          & 1.429          & 1.637          & 1.857          & 2.083          & 5.664          & 5.993          & 6.949          & 8.23           & 9.613          \\
PCCN~\cite{beetz2023multi}              & 1.628          & 1.746          & 1.992          & 2.323          & 2.731          & 6.723          & 7.065          & 8.277          & 9.761          & 11.662         \\
\emph{HeartFormer (Ours)}         & \textbf{1.194} & \textbf{1.373} & \textbf{1.585} & \textbf{1.799} & \textbf{2.037} & \textbf{5.417} & \textbf{5.678} & \textbf{6.285} & \textbf{7.401} & \textbf{8.662} \\ \midrule
\end{tabular}
}
\label{tab2}
\end{table*}

\subsection{Evaluation on HeartCompv1 dataset}
\label{sec:Evaluation_HeartCompv1}

We compare our proposed HeartFormer with SOTA single-class point cloud completion methods, including FoldingNet~\cite{yang2017foldingnet}, PCN~\cite{yuan2018pcn}, SnowflakeNet~\cite{xiang2022snowflake}, PoinTr~\cite{yu2021pointr}, FBNet~\cite{yan2022fbnet}, AdaPoinTr~\cite{yu2301adapointr}, AnchorFormer~\cite{chen2023anchorformer}, CRA-PCN~\cite{rong2024cra}, ODGNet~\cite{cai2024orthogonal}, PointAttN~\cite{wang2024pointattn}, and SymmCompletion~\cite{yan2025symmcompletion}, as well as the multi-class completion method PCCN~\cite{beetz2023multi}. We evaluate all methods on six test sets exhibiting different degrees of misalignment: none, mild, medium, strong, severe, and a mixed setting that combines all levels. These configurations simulate realistic inter-slice motion artifacts commonly observed in dynamic cardiac MRI, enabling a systematic assessment of robustness of each model under varying conditions. 

\begin{figure}[!t]
\centering
\includegraphics[width=\linewidth]{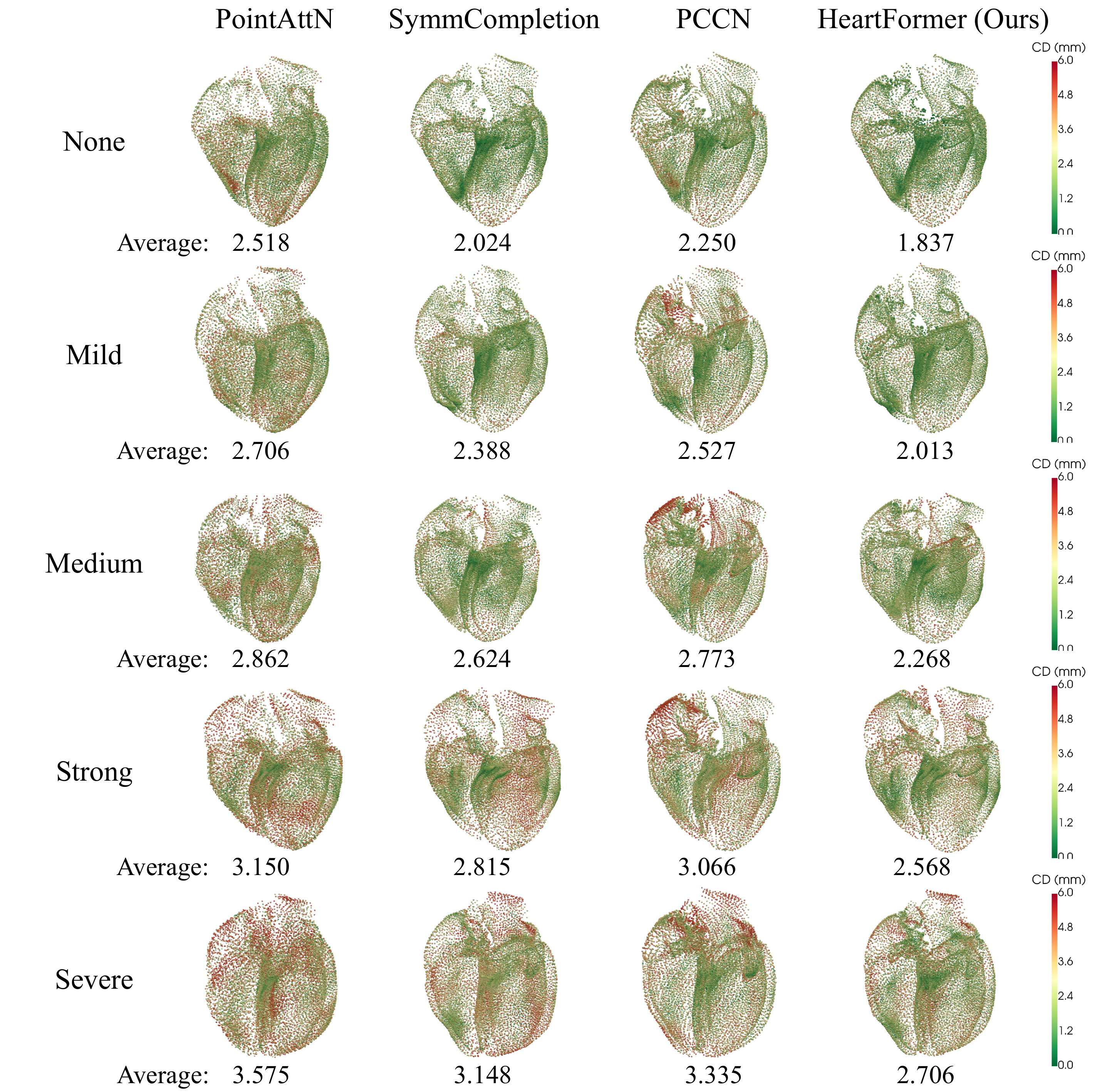} 
\caption{\textbf{Comparison of reconstruction results under five different levels of misalignment in the HeartCompv1 dataset.} The point colors visualize the Chamfer Distance between the reconstructed point cloud and the corresponding ground truth.}
\label{fig_3}
\end{figure}

Table~\ref{tab1_1} presents the quantitative results on the mixed-misalignment setting of the HeartCompv1 dataset, along with comparisons of model parameters and FLOPs. The proposed HeartFormer consistently achieves the best performance across all metrics compared with both SOTA single-class and multi-class completion methods, while requiring fewer parameters and lower computational cost.

\begin{table}[]
\caption{\textbf{Quantitative comparison of different models on the UK Biobank.} HeartFormer significantly outperforms PCCN across all six cardiac substructures under every evaluation metric.}
\centering
\resizebox{0.9\linewidth}{!}{
\begin{tabular}{l|cc|cc}
\midrule
\multirow{2}{*}{Substructure} & \multicolumn{2}{c|}{CD $\downarrow$} & \multicolumn{2}{c}{HD $\downarrow$} \\
                              & PCCN~\cite{beetz2023multi}    & HeartFormer        & PCCN~\cite{beetz2023multi}    & HeartFormer       \\ \midrule
LV Endo                       & 2.712   & \textbf{1.382}   & 12.423  & \textbf{11.184}  \\
LV Epi                        & 3.460   & \textbf{1.551}   & 15.485  & \textbf{12.057}  \\
RV Endo                       & 3.360   & \textbf{1.790}    & 16.550  & \textbf{15.924} \\
RV Epi                        & 3.374   & \textbf{1.755}   & 17.706  & \textbf{11.456}  \\
LA                            & 5.036   & \textbf{2.344}   & 20.398  & \textbf{19.749}  \\
RA                            & 3.723   & \textbf{2.037}   & 16.217  & \textbf{15.038} \\ \midrule
\end{tabular}
}
\label{tab4}
\end{table}

\begin{figure}[!t]
\small
\centering
\includegraphics[width=3in]{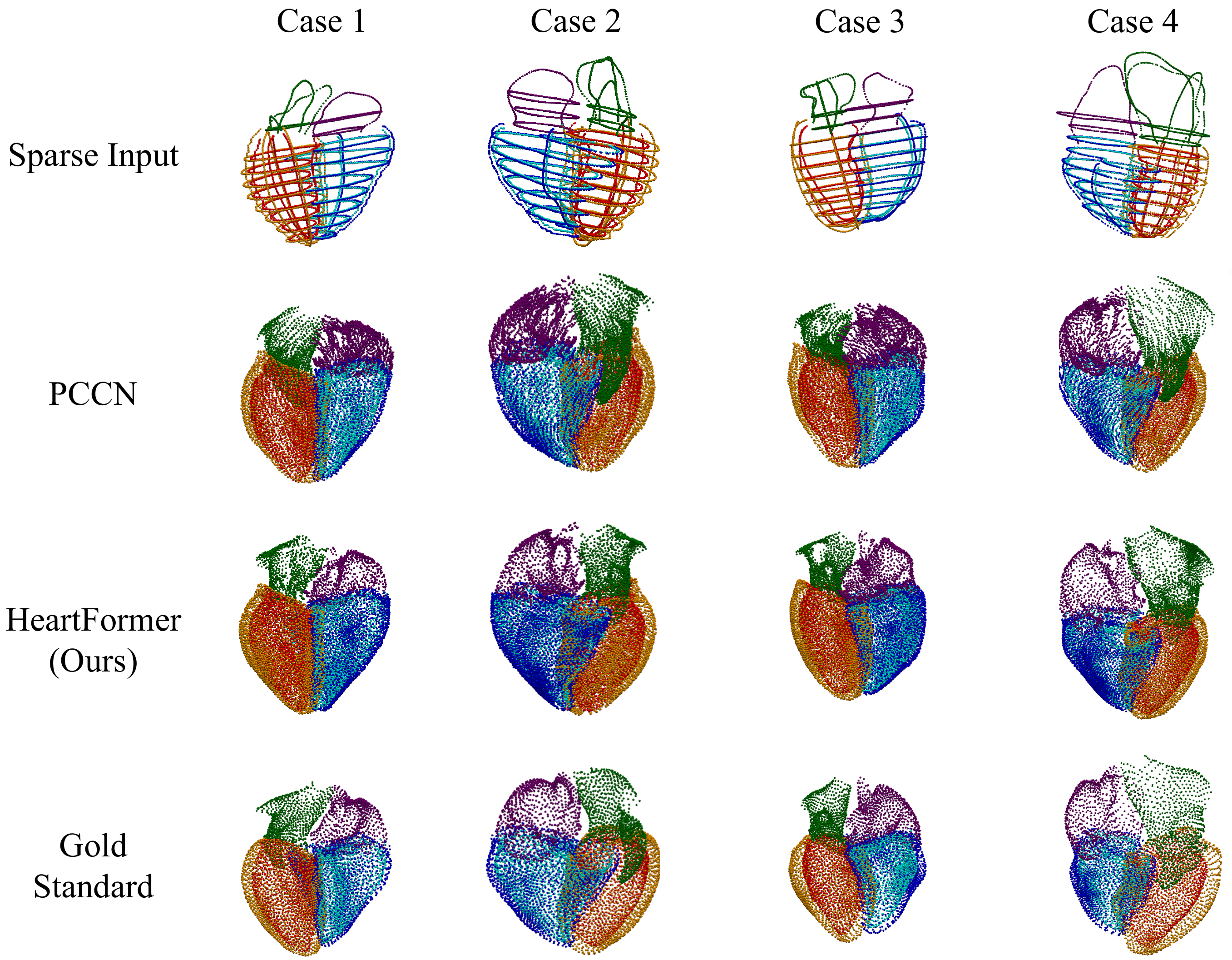} 
\caption{\textbf{Comparison of reconstruction results for four subjects from the UK Biobank dataset.} The reconstructions produced by HeartFormer more closely resemble the gold standard than those generated by PCCN.}
\label{fig_5}
\end{figure}

Figure~\ref{fig_2} and \cref{tab2} present qualitative and quantitative comparisons across varying misalignment levels. Single-class models tend to reconstruct the four-chamber heart surface as a single entity, lacking anatomical differentiation. In contrast, multi-class models generate structure-aware outputs for each cardiac chamber. As shown in the visual results, PCCN often produces sparse or line-shaped point clouds with uneven distributions, whereas HeartFormer yields dense and anatomically consistent surfaces across all settings. The advantage of HeartFormer becomes more evident as the degree of misalignment increases, demonstrating stronger robustness. For quantitative comparison in \cref{tab2}, our method achieves the best performance across all misalignment levels and metrics. For the medium misalignment setting, which most closely resembles real clinical conditions, HeartFormer maintains CD result below 1.6 mm, further highlighting its effectiveness for realistic cardiac shape reconstruction, particularly under challenging misalignment scenarios. 

Figure~\ref{fig_3} presents qualitative and quantitative comparisons across five levels of misalignment. The proposed HeartFormer effectively handles spatial misalignment and achieves superior performance compared to both single-class and multi-class completion methods. These results demonstrate the robustness of HeartFormer in both quantitative and qualitative evaluations, particularly under non-ideal conditions commonly encountered in clinical data. Additional results are provided in the Supplementary.




\begin{table}[]
\caption{\textbf{Estimated 3D volumetric clinical metrics at end-diastole.} Values are reported as mean ± standard deviation. To assess the clinical relevance of the reconstructed point clouds, standard volumetric indices are computed at end-diastole.}
\centering
\resizebox{\linewidth}{!}{
\begin{tabular}{c|c|c|c}
\midrule
Sex                     & Clinical Metric & Proposed     & Benchmark \\ \midrule
\multirow{4}{*}{Female} & LVV (ml)        & 130.99 ± 11.38 & 124 ± 21  \\
                        & RVV (ml)        & 142.28 ± 10.75 & 130 ± 24  \\
                        & LAV (ml)        & 57.52 ± 10.42   & 62 ± 17   \\
                        & RAV (ml)        & 62.77 ± 9.82   & 69 ± 17   \\ \midrule
\multirow{4}{*}{Male}   & LVV (ml)        & 159.78 ± 19.26 & 166 ± 32  \\
                        & RVV (ml)        & 185.19 ± 20.03 & 182 ± 36  \\
                        & LAV (ml)        & 66.16 ± 12.28   & 71 ± 19   \\
                        & RAV (ml)        & 92.82 ± 25.93  & 93 ± 27   \\ \midrule
\end{tabular}
}
\label{tab5}
\end{table}

\subsection{Analysis on UK Biobank data}
\label{sec:Evaluation_UKBiobank}

Table~\ref{tab4} presents quantitative comparisons for each cardiac substructure on the UK Biobank dataset. The proposed HeartFormer consistently outperforms PCCN, demonstrating markedly better performance in real clinical settings. For the atrial substructures, reconstruction remains challenging due to the limited and sparse atrial point clouds in the input data. While PCCN struggles to recover meaningful atrial geometry, HeartFormer still achieves reliable and competitive scores, highlighting its superior generalization.

\begin{figure}[!t]
\centering
\small
\includegraphics[width=\linewidth]{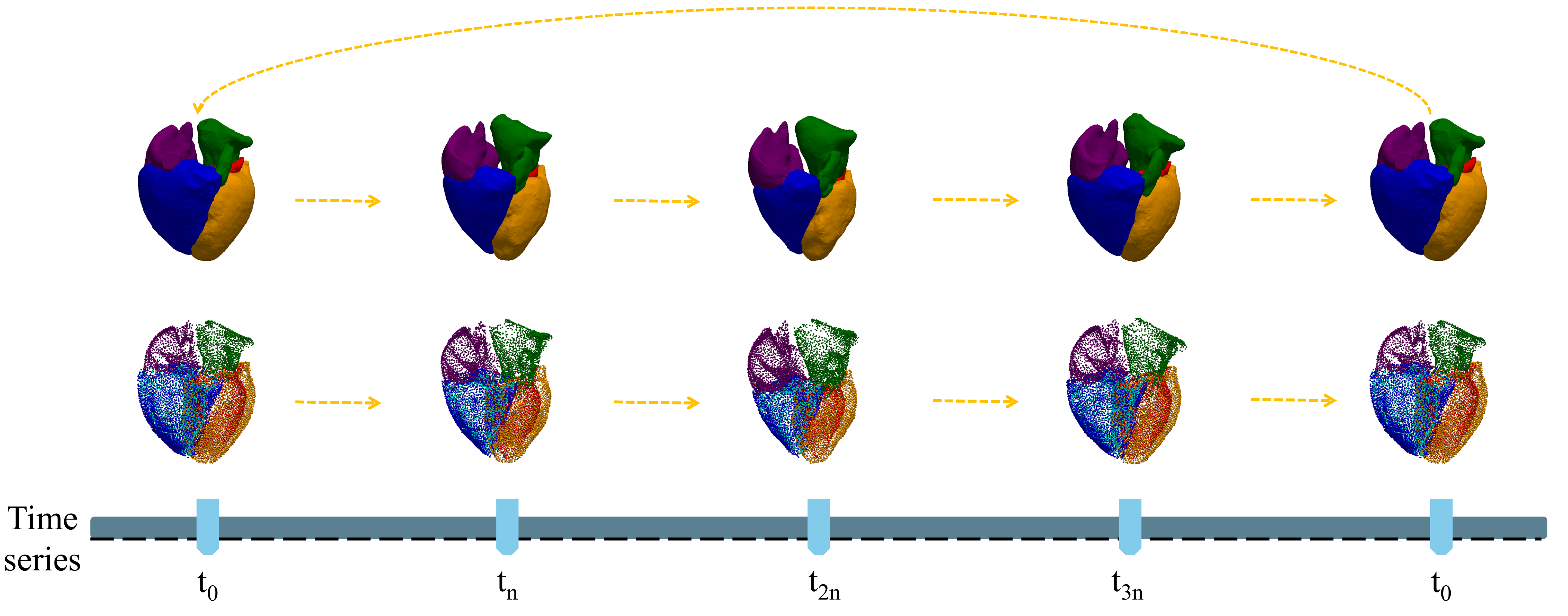} 
\caption{\textbf{4D cardiac point cloud and mesh reconstruction for a UK Biobank subject.} The dynamic video is available on GitHub.}
\label{fig_7}
\end{figure}

Figure~\ref{fig_5} presents reconstruction results for four representative subjects from the UK Biobank dataset. The proposed HeartFormer produces more anatomically faithful and complete cardiac surfaces compared to PCCN. Table~\ref{tab5} summarizes the estimated end-diastolic 3D volumetric clinical metrics for female and male subjects, alongside the reference values from~\cite{petersen2016reference}. Our 3D reconstructions achieve physiologically plausible scores across all metrics and accurately capture sex-related differences. \cref{fig_7} shows the 4D cardiac point cloud and mesh reconstruction for a UK Biobank subject, effectively capturing the dynamic motion of cardiac contraction and relaxation. Both qualitative and quantitative analyses confirm that HeartFormer delivers accurate, robust, and clinically meaningful reconstruction performance. More results can be found in the Supplementary.

\subsection{Ablation Studies}

We conduct extensive ablation studies to evaluate our design choices, reporting both qualitative and quantitative results on the HeartCompv1 dataset. Additional ablation analyses are provided in the Supplementary.

\noindent\textbf{Are global and substructure geometric points and features all essential for model performance?}
To evaluate the impact of global and substructure geometric representations, we conduct an ablation study by removing them from our framework along with their associated generation and processing modules. As shown in \cref{tab6}, both global and substructure geometric points and features substantially affect model performance. In particular, the substructure geometric points and features contribute more significantly than the global geometric components.

\begin{table}[]
\centering
\caption{\textbf{Quantitative comparison of the effects of global and substructure geometric points and features.}}
\begin{tabular}{cc|cc|cc}
\midrule
\multicolumn{2}{c|}{SA-GSA} & \multicolumn{2}{c|}{SA-SSA} & \multirow{2}{*}{CD $\downarrow$} & \multirow{2}{*}{SSD $\downarrow$} \\
$F_{glo}$         & $P_{glo}$         & $F_{sub}$         & $P_{sub}$         &                        &                        \\ \midrule
\cmark            & \cmark            &              &              & 1.701                  & 1.769                  \\
\cmark            & \cmark            &              & \cmark            & 1.678                  & 1.741                  \\
             &              & \cmark            & \cmark            & 1.689                  & 1.759                  \\
             & \cmark            & \cmark            & \cmark            & 1.648                  & 1.711                  \\
\cmark            & \cmark            & \cmark            & \cmark            & \textbf{1.596}         & \textbf{1.661}         \\ \midrule
\end{tabular}
\label{tab6}
\end{table}



\begin{table}[]
\centering 
\caption{\textbf{Quantitative comparison of loss functions.} SA-CD represents the Semantic-Aware Geometry loss, with CD computed at coarse, middle, and fine stages.}
\resizebox{\linewidth}{!}{
\begin{tabular}{ccc|c|cc}
\midrule
\multicolumn{3}{c|}{SA-CD} & \multirow{2}{*}{Geometry loss} & \multirow{2}{*}{CD $\downarrow$} & \multirow{2}{*}{SSD $\downarrow$} \\
$L_{coarse}$   & $L_{mid}$   & $L_{fine}$  &                                &                        &                        \\ \midrule
          &        &        & \cmark                              & \multicolumn{2}{c}{Worng labeling}              \\
\cmark         &        & \cmark      &                                & 1.624                  & 1.692                   \\
          & \cmark      & \cmark      &                                & 1.637                  & 1.707                  \\
\cmark         & \cmark      & \cmark      &                                & \textbf{1.596}         & \textbf{1.661}         \\
\cmark         & \cmark      & \cmark      & \cmark                              & 1.616                  & 1.685                  \\ \midrule
\end{tabular}
}
\label{tab7}
\end{table}

\noindent\textbf{Are all components of the Semantic-Aware Geometry loss essential, and is the geometry loss beneficial?}
\Cref{tab7} presents an ablation of the loss function. Removing any component in Semantic-Aware Geometry loss leads to a consistent drop in performance, confirming the necessity of multi-scale semantic alignment. Models trained without Semantic-Aware guidance show clear semantic mislabeling, while the additional geometry loss does not yield further improvement, suggesting that the Semantic-Aware Geometry loss already provides strong geometric regularization. These findings highlight that accurate semantic correspondence across scales plays a pivotal role for achieving robust and coherent reconstruction quality.
\section{Conclusions}
\label{sec:Conclusions}

In this work, we present a novel geometric deep learning framework for 3D four-chamber cardiac reconstruction from cine MRI using multi-class point cloud completion. Central to our framework is \textit{HeartFormer}, a novel completion network that jointly leverages global anatomical context and local substructural priors, while adaptively refining substructure representations under semantic-conditioned guidance to achieve high-fidelity cardiac reconstruction. We also introduce \textit{HeartCompv1}, the first large-scale dataset for multi-class cardiac point cloud completion. We hope HeartFormer and HeartCompv1 will inspire future research in this direction and contribute to advancing clinically impactful 3D and 4D cardiac modeling.

\section*{Acknowledgements}
The authors would like to acknowledge the use of the facilities and services of the Institute of Biomedical Engineering (IBME), Department of Engineering Science, University of Oxford, in conducting this research.
The work of A. Banerjee was supported by the Royal Society University Research Fellowship (grant no. URF\textbackslash R1\textbackslash 221314) and the BHF Oxford Centre of Research Excellence (BHF RE/24/130024).  
This research has been conducted using the UK Biobank Resource under Application Number ‘135228’.


\small
    \bibliographystyle{unsrt} 
    \bibliography{main}
%
\appendix
\clearpage
\setcounter{page}{1}
\renewcommand{\thefigure}{S\arabic{figure}}
\setcounter{figure}{0}
\renewcommand{\thetable}{S\arabic{table}}
\setcounter{table}{0}

\renewcommand{\theequation}{S\arabic{equation}}
\setcounter{equation}{0}

\maketitlesupplementary


Our supplementary materials provide an overview of the entire workflow (Sec.~\ref{sec:Workflow}), detailed implementation descriptions (Sec.~\ref{sec:Implementation}), extended experimental analyses that complement the findings reported in the main paper (Sec.~\ref{sec:Additional_Experimental_Analyses}), and additional ablation studies (Sec.~\ref{sec:Additional_Ablation_Studies}). We further include introductions to the released code and the discussion section in Sec.~\ref{sec:Code} and Sec.~\ref{sec:Discussion}, respectively. We also include three representative examples of generated 4D cardiac point cloud videos and a video visualizing the point cloud completion training process in the supplementary material, along with the code.

\section{Workflow}
\label{sec:Workflow}

\begin{figure*}[!t]
\centering
\includegraphics[width=0.9\linewidth]{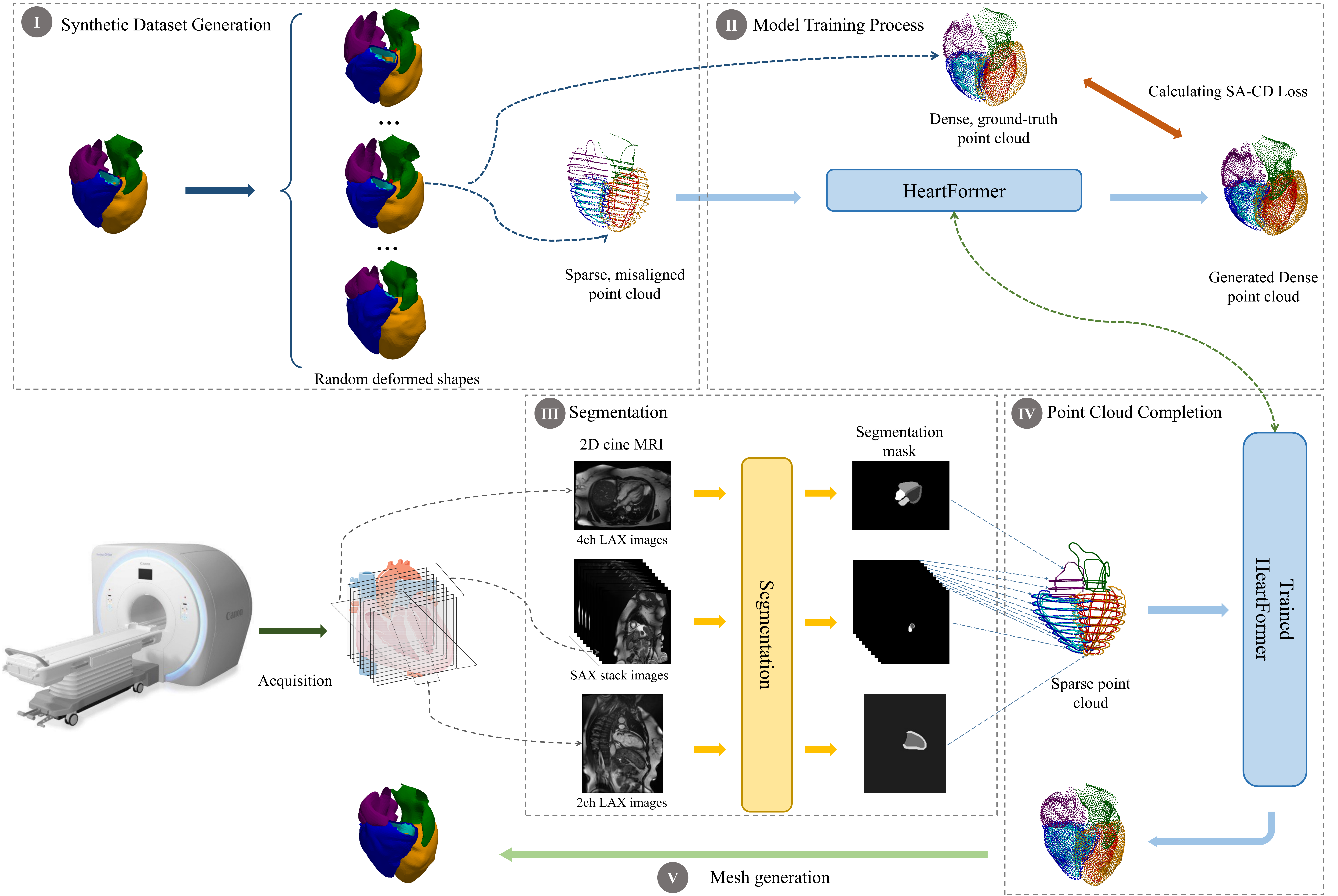} 
\caption{\textbf{Overview of the entire workflow.} The overall workflow comprises of three major components: (I) Synthetic Dataset Generation, (II) Model Training Process, and the Reconstruction Pipeline, which further consists of (III) Segmentation, (IV) HeartFormer Point Cloud Completion, and (V) Mesh Generation.
}
\label{fig_workflow}
\end{figure*}

The overall workflow consists of three major components: (I) Synthetic Dataset Generation, (II) Model Training Process, and the Reconstruction Pipeline, which involves (III) Segmentation, (IV) HeartFormer Point Cloud Completion, and (V) Mesh Generation, as shown in Fig.~\ref{fig_workflow}.

\subsection{Synthetic Dataset Generation}

\label{sec:SSM}

We generate a synthetic dataset for 3D cardiac point cloud reconstruction, based on the Full Heart principal component analysis (PCA) model provided in \cite{hoogendoorn2012high}.

For each sample, we obtain a sparse, misaligned point cloud, a dense, ground-truth point cloud, and a high-resolution mesh. We introduce \textit{HeartCompv1}, the first publicly available large-scale dataset containing 17,000 high-resolution 3D cardiac models, provided in both mesh and point cloud formats.

\subsection{Model Training Process}

The training set comprises $N = 10{,}000$ samples, while both the validation set and each of the six test sets contain $N = 1{,}000$ samples. For every sample, the ground-truth (GT) point cloud is uniformly resampled to 16,384 points through random sampling, and the corresponding sparse input is resampled to 7,500 points. Model optimization is conducted using the Semantic-Aware Chamfer Distance (SA-CD) loss described in Sec.~\ref{sec:loss}.

\subsection{Segmentation}

The segmentation stage processes the SAX, 4CH LAX, and 2CH LAX cine-MRI slices from the UK Biobank dataset. For the SAX stack and 4CH LAX slices, we adopt the approach of Bai et al.~\cite{bai2018automated}, which achieves human-level accuracy in cardiac structure segmentation from cine MRI. For the 2CH LAX view, we utilize the pre-trained model provided by Beetz et al.~\cite{beetz2023multi}. Atrial contours are manually annotated using the tools described in Banerjee et al.~\cite{banerjee2021completely}, from which segmentation masks are subsequently generated.

\subsection{HearFormer Point Cloud Completion}

In the main paper, we provided a detailed introduction to the overall architecture of HeartFormer and its components, SA-DSTNet and SA-GFRTNet, in Sec.~\ref{sec:HeartFormer}. In this subsection, we present the detailed configurations of both GSBlock and SSBlock, as mentioned in Sec.~\ref{sec:SA-DSTNet}.

\begin{figure}[!t]
\centering
\includegraphics[width=\linewidth]{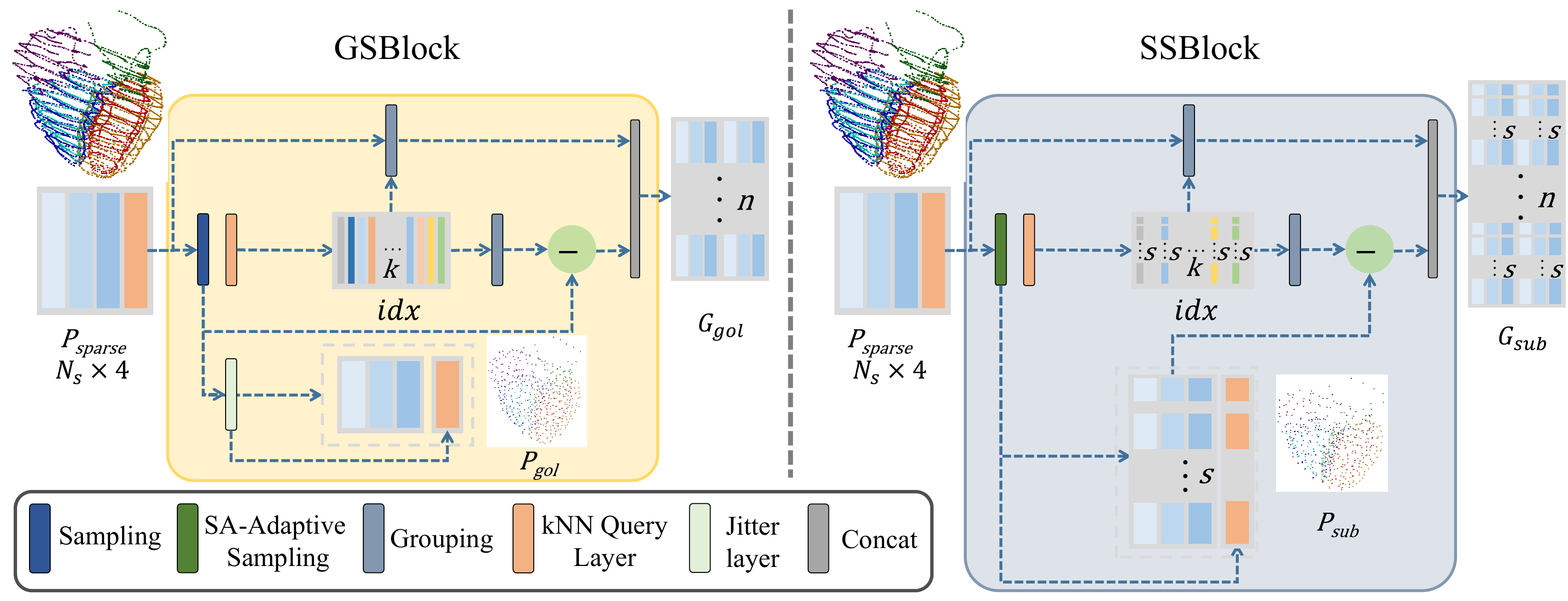} 
\caption{The architecture of GSBlock and SSBlock.}
\label{fig_block}
\end{figure}

\subsubsection{Global Sampling Block (GSBlock).} 
To efficiently capture both global structures and local geometric relations in large-scale point clouds, we propose the GSBlock, as shown in Fig.~\ref{fig_block}. Unlike conventional set abstraction layers that rely solely on local sampling and neighborhood grouping, GSBlock introduces a hybrid sampling-and-grouping mechanism that integrates global-aware sampling with label-consistent feature aggregation.

Given an input point cloud with its label embeddings $P_{sparse} = \{(x_i, y_i, z_i, l_i)\}_{i=1}^{N_s}$, we first perform Furthest Point Sampling (FPS) to select a subset of $n_p$ representative centers:
\begin{equation}
\mathcal{C} = \operatorname{FPS}(P_{sparse}, n), \quad
\mathbf{p}_j = (x_j, y_j, z_j, l_j) \in \mathcal{C}.
\end{equation}
\begin{equation}
P_{gol} = \mathcal{J}(\mathcal{C}).
\end{equation}

$\mathcal{J}$ represents the function of Jitter layer.


For each centroid $\mathbf{p}_j$, GSBlock gathers its $k$ nearest neighbors from the input space using the $k$-Nearest Neighbor (kNN) operator:
\begin{equation}
\mathcal{N}_j = \operatorname{kNN}(\mathbf{p}_j, P_{sparse}, k),
\end{equation}
where $\mathcal{N}_j = \{\mathbf{p}_{j,1}, \mathbf{p}_{j,2}, \ldots, \mathbf{p}_{j,k}\}$ denotes the indices of the neighboring points. Each local region is translated into a centroid-aligned coordinate frame by
\begin{equation}
\Delta \mathbf{p}_{j,i} = \mathbf{p}_{j,i} - \mathbf{p}_j,
\end{equation}
which improves local geometric invariance and learning stability. To ensure semantic consistency, GSBlock aggregates both geometric and label-aware features:


\begin{equation}
\mathbf{g}_{j,i}^g = [\, \Delta \mathbf{p}_{j,i},\ \mathbf{p}_{j,i} \,],
\end{equation}
where $[\cdot]$ denotes feature concatenation. The grouped tensor is thus
\begin{equation}
G_{gol} = \{ \mathbf{g}_{j,i}^g \}_{j=1,i=1}^{n,k}. 
\end{equation}
This representation jointly encodes local geometry, global position, and label semantics, serving as a rich descriptor for downstream feature extraction.


The GSBlock effectively maintains global uniformity while preserving fine-grained local structures. By combining FPS-based global sampling with label-aware kNN grouping, GSBlock captures multi-scale contextual cues across complex and heterogeneous point distributions. In addition, the jitter layer enhances model robustness by simulating the annotation variability often observed among radiologists in clinical settings.

\subsubsection{Substructure Sampling Block (SSBlock).} 
SSBlock extends the GSBlock with an adaptive class-balanced sampling strategy and a subsequent kNN-based grouping mechanism, jointly designed to enhance both geometric uniformity and category-level fairness in feature learning, as illustrated in Fig.~\ref{fig_block}. SSBlock dynamically allocates sampling quotas across classes. Let the total number of sampled points be $N_s$, and $\alpha$ be a balance coefficient controlling the emphasis on rare categories.  
We first compute the adaptive sampling ratio for class $c$ as:
\begin{equation}
r_c = \frac{n_c^{-\alpha}}{\sum_{j=1}^{C} n_j^{-\alpha}},
\end{equation}
where $n_c$ is the number of points belonging to class $c$.  
The number of points to sample from each class is:
\begin{equation}
n_c^{\ast} = \lfloor r_c \cdot N_s \rfloor.
\end{equation}
This formulation adaptively boosts the contribution of sparse classes when $\alpha > 0$, effectively mitigating bias from dense categories.

For each class $c$, we then perform FPS within its subset:
\begin{equation}
\mathcal{C}_c = \operatorname{FPS}(\mathbf{P}_c, n_c^{\ast}),
\end{equation}
ensuring that the sampled centroids $\mathcal{C}_c = \{\mathbf{p}_{c,1}, \ldots, \mathbf{p}_{c,n_c^{\ast}}\}$ are spatially well distributed within their own class region.  
The final sampled set is the union over all classes:
\begin{equation}
\mathcal{C} = \bigcup_{c=1}^{C} \mathcal{C}_c, \quad |\mathcal{C}| = N_s.
\end{equation}

Around each centroid $\mathbf{p}_j \in \mathcal{C}$, we perform a $k$-Nearest Neighbor search over the entire input cloud:
\begin{equation}
\mathcal{N}_j = \operatorname{kNN}(\mathbf{p}_j, P_{sparse}, k),
\end{equation}
yielding $k$ local neighbors.  
Each local region is translated to a centroid-aligned frame:
\begin{equation}
\Delta \mathbf{p}_{j,i} = \mathbf{p}_{j,i} - \mathbf{p}_j,
\end{equation}
and its label-aligned feature vector is defined as:
\begin{equation}
\mathbf{g}_{j,i}^s = [\, \Delta \mathbf{p}_{j,i},\ \mathbf{p}_{j,i} \,].
\end{equation}

Finally, the grouped tensor is represented as:
\begin{equation}
\mathbf{G}_{sub} = \{ \mathbf{g}_{j,i}^s \}_{j=1,i=1}^{n,k}.
\end{equation}

Through adaptive sampling and label-consistent grouping, SSBlock achieves balanced class representation while preserving geometric fidelity. The dynamic interplay between the adaptive sampling ratio $r_c$ and the global–local neighborhood construction enables the network to learn robust, class-invariant representations. This design substantially enhances recognition and reconstruction performance, particularly for minority categories in the imbalanced HeartCompv1 dataset.

\subsection{Mesh Generation}

We generate high-fidelity surface meshes from the reconstructed point clouds through a two-stage meshing pipeline that integrates Poisson reconstruction and Ball Pivoting~\cite{bernardini2002ball}. Surface normals are first estimated and globally oriented to ensure geometric consistency. A preliminary mesh is obtained via Poisson reconstruction at a moderate octree depth, followed by Poisson disk resampling to achieve a uniform point distribution. The resampled points are then processed by the Ball Pivoting algorithm using multi-scale radii derived from the average point spacing. To ensure watertightness and structural integrity, topological artifacts and non-manifold elements are subsequently repaired with MeshFix~\cite{attene2010lightweight}, resulting in clean triangular meshes suitable for further analysis.

\subsection{Dataset}
\label{sec:dataset}

The training dataset consists of $N = 10{,}000$ samples with varying degrees of misalignment. Both the validation set and each of the six test sets contain $N = 1{,}000$ samples. The validation set also includes samples exhibiting different levels of misalignment. Each sample comprises a sparse, misaligned input point cloud of the heart and a high-resolution ground-truth point cloud annotated with six anatomical structures: LV endocardium, LV epicardium, RV endocardium, RV epicardium, LA, and RA. For training, validation, and testing, each sample contains a resampled sparse input point cloud with 7,500 points and a resampled dense GT point cloud with 16,384 points.

\section{Implementation Details}
\label{sec:Implementation}

\subsection{HeartFormer}

\subsubsection {Network Architecture}

Our proposed point cloud completion model, HeartFormer, is designed to output three levels of prediction: a coarse point cloud of shape $[B \times 1024 \times 4]$, middle point cloud of shape $[B \times 2048 \times 4]$ and a fine point cloud of shape $[B \times 16384 \times 4]$, where $B$ is the batch size.

The CardioNet architecture is configured with upsampling factors of 2 and 8, with a coarse upsampling factor of 2. Details of the architecture are described in the main paper.

\subsubsection{Training Protocol}
\label{sec:Training Protocol}

HeartFormer is trained for 420 epochs with a batch size of 8 on an NVIDIA Tesla V100 GPU. The model is optimized using AdamW with an initial learning rate of $2\times10^{-4}$ and a weight decay of $5\times10^{-4}$. A WarmUpCosLR scheduler is employed to linearly increase the learning rate during the first 20 warm-up epochs, followed by a cosine annealing schedule that decays it from $2\times10^{-4}$ to $1\times10^{-5}$ over the remaining epochs.

During training, we monitor both training and validation losses per epoch and visualize randomly selected validation samples, including the input, coarse prediction, dense prediction, and ground truth point clouds (Fig.~\ref{fig_training}). Model checkpoints are saved whenever the validation loss reaches a new minimum for either the coarse or fine outputs. The best-performing models, based on coarse and dense Chamfer Distance (CD) losses, are stored separately, and the model with the lowest dense CD loss is used for final evaluation.

\subsubsection {Evaluation and Logging}
\label{sec:Evaluation and Logging}
\begin{figure}[!h]
    \centering
    \includegraphics[width=3.5in]{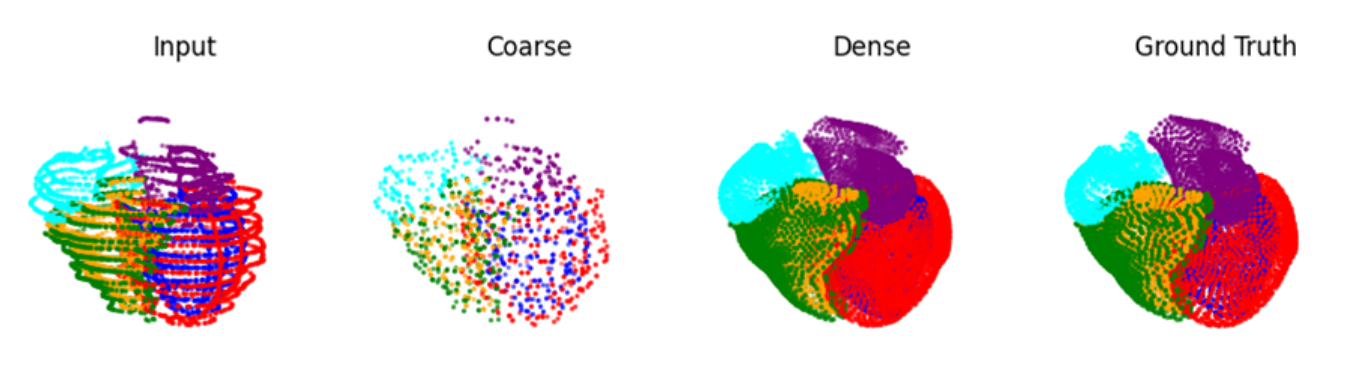} 
    \caption{Visualization of the training performance. From left to right: input point cloud, coarse prediction, dense prediction, and ground truth.} 
    \label{fig_training}
    \end{figure}

We present a subset of validation samples in Fig.~\ref{fig_training}, which is saved at each epoch, providing qualitative insight into model performance.


\subsection {FoldingNet~\cite{yang2017foldingnet}}



\subsubsection{Network Architecture} 
FoldingNet employs a point cloud encoder-decoder architecture. The encoder extracts global features with an output channel dimension of 1,024. The decoder reconstructs a dense point cloud of 16,384 points using a two-stage folding-based deformation mechanism applied to a 2D grid.

Our implementation of FoldingNet is based on the official code provided in \url{https://github.com/qinglew/FoldingNet}

\subsubsection{Training Protocol} 
We use the Adam optimizer with an initial learning rate of 0.0001 and no weight decay (i.e., the L2 regularization coefficient is set to zero). A StepLR scheduler is applied, reducing the learning rate by a factor of 0.5 every 50 epochs.

We use the same parameter settings as stated in the respective studies to ensure comparability. The model is trained for 1000 epochs with a batch size of 8 and gradient accumulation step of 1. Chamfer Distance is used as loss function.

\subsection {PCN~\cite{yuan2018pcn}}



\subsubsection{Network Architecture} 
PCN is designed with a coarse-to-fine decoder structure. It first generates a coarse point cloud and then refines it using a folding-based deformation mechanism. The encoder outputs a global feature vector of dimension 1,024, and the final output contains 16,384 points.

Our implementation of PCN is based on the official code provided in \url{https://github.com/qinglew/PCN-PyTorch}

\subsubsection{Training Protocol} 
The model is trained for 1000 epochs with a batch size of 8 and gradient accumulation step of 1. Chamfer Distance is used as loss function.

We use the same parameter settings as stated in the PCN studies to ensure comparability. We use the Adam optimizer with a learning rate of 0.0001 and no weight decay. The learning rate follows a \texttt{StepLR} schedule with a decay factor of 0.5 every 50 epochs.

\subsection {SnowflakeNet~\cite{xiang2022snowflake}}


\subsubsection{Network Architecture} 
The network begins with a coarse point cloud of 512 points and progressively refines it to finer resolutions. The feature embedding dimension is set to 512, with an intermediate prediction size of 256 points. Upsampling is performed in two stages with factors [4, 8]. A neighborhood search radius of 2 is used during local refinement.

Our implementation of SnowflakeNet is based on the official code provided in 
\url{https://github.com/AllenXiangX/SnowflakeNet}

\subsubsection{Training Protocol} 
We use the Adam optimizer with an initial learning rate of 0.001 and no weight decay. The learning rate is scheduled using a GradualWarmup strategy: the learning rate is linearly increased to the base value over the first 200 epochs, followed by a step decay scheme that reduces the learning rate by a factor of 0.5 every 50 epochs.

We use the same parameter settings as stated in the SnowFlakeNet studies to ensure comparability. The model is trained for 420 epochs with a batch size of 8 and a gradient accumulation step of 1. Chamfer Distance is used as loss function.

\subsection{PoinTr \cite{yu2021pointr}}

\subsubsection{Network Architecture}
PoinTr adopts a Transformer-based encoder–decoder framework that predicts dense point clouds of 16,384 points. The model employs a single kNN layer to capture local geometric structures, and an embedding dimension of 384 for feature representation. 
This architecture enables effective global–local feature interaction for reconstructing fine-grained 3D geometry.

Our implementation of PoinTr is based on the official code provided in \url{https://github.com/yuxumin/PoinTr}

\subsubsection{Training Protocol}
PoinTr is trained for 420 epochs with an effective batch size of 8 and a gradient accumulation step of 1. 
The AdamW optimizer is used with an initial learning rate of $5\times10^{-4}$ and a weight decay of $5\times10^{-4}$. 
A LambdaLR scheduler decays the learning rate by a factor of 0.9 every 21 epochs, with a minimum learning rate 
set to 2\% of the initial value. Additionally, a batch normalization momentum scheduler (BNM scheduler) 
reduces the momentum by a factor of 0.5 every 21 epochs, starting from 0.9 and capped at a lower bound of 0.01. The model is optimized using the Chamfer Distance as the loss function.

\subsection{FBNet \cite{yan2022fbnet}}

\subsubsection{Network Architecture}
FBNet adopts a hierarchical upsampling framework designed to progressively reconstruct dense point clouds 
from sparse inputs. The model is configured with upsampling factors of $1$, $2$, and $16$ across three 
refinement stages, with each cycle repeated three times. 
The network takes $512$ input points and generates $128$ coarse point features 
before producing the final dense output. 

Our implementation of FBNet is based on the official code provided in \url{https://github.com/hikvision-research/3DVision}

\subsubsection{Training Protocol}
FBNet is trained for 420 epochs with an effective batch size of 8 and a gradient accumulation step of 1. The AdamW optimizer is used with an initial learning rate of $2\times10^{-4}$ and a weight decay of $5\times10^{-4}$. A WarmUpCosLR scheduler linearly increases the learning rate during the first 20 warm-up epochs, followed by a cosine annealing schedule that decays the learning rate from $2\times10^{-4}$ to $1\times10^{-5}$ over the remaining epochs. The model is optimized using the Chamfer Distance as the loss function.

\subsection {AdaPoinTr~\cite{10232862}}

\subsubsection{Network Architecture} 
The model follows a Transformer-style encoder-decoder framework. It generates dense point clouds of 16,384 points from 512 query tokens. The encoder consists of 6 layers with an embedding dimension of 384, along with 6 attention heads, and 2 graph groups with a neighborhood size $k=8$. The attention block style starts with a graph-attention block followed by five self-attention blocks. Feature aggregation is performed via concatenation.

The decoder contains 8 layers with similar configurations. Both self-attention and cross-attention branches adopt the same block styling: one graph-attention layer followed by seven layers. Concatenation is used to combine features across layers.

Our implementation of AdaPoinTr is based on the official code provided in \url{https://github.com/yuxumin/PoinTr}

\subsubsection{Training Protocol} 

AdaPoinTr adopts the AdamW optimizer with an initial learning rate of 0.0001 and a weight decay of 0.0005. A LambdaLR scheduler decays the learning rate by a factor of 0.9 every 21 epochs, with a minimum learning rate set to 2\% of the initial value. Additionally, a batch normalization momentum scheduler decays the momentum by 0.5 every 21 epochs, starting from 0.9 and capped at a lower bound of 0.01.

We use the same parameter settings as stated in the AdaPoinTr studies to ensure comparability. The model is trained for 420 epochs with a batch size of 8 and gradient accumulation step of 1. Chamfer Distance is used as loss function.

\subsection{AnchorFormer \cite{chen2023anchorformer}}

\subsubsection{Network Architecture}
AnchorFormer follows a Transformer-based encoder–decoder framework that progressively refines point cloud representations. 
The model predicts dense point clouds of 16,384 points from 256 query tokens. 
The encoder consists of eight layers with an embedding dimension of 384, while the decoder contains six layers of similar configuration. 
Both sparse and dense branches are controlled by two expansion factors, 
$\lambda_{\text{sparse}} = 0.5$ and $\lambda_{\text{dense}} = 1.2$, 
to balance the generation of coarse and fine structures. 
The loss design incorporates four components with equal weighting: 
sparse loss, dense loss, sparse penalty, and dense penalty, each assigned a weight of 1.0.

Our implementation of AnchorFormer is based on the official code provided in \url{https://github.com/chenzhik/AnchorFormer}

\subsubsection{Training Protocol}
AnchorFormer is trained for 420 epochs with an effective batch size of 8 and a gradient accumulation step of 1. 
The AdamW optimizer is adopted with an initial learning rate of $2\times10^{-4}$ and a weight decay of $5\times10^{-4}$. 
A WarmUpCosLR scheduler linearly increases the learning rate during the first 20 warm-up epochs 
and then applies a cosine annealing schedule that decays the learning rate from $2\times10^{-4}$ to $1\times10^{-5}$ 
over the remaining epochs. The model is optimized using the Chamfer Distance as the primary loss metric.

\subsection{CRA-PCN \cite{rong2024cra}}

\subsubsection{Network Architecture}
CRA-PCN follows a completion-based architecture designed to reconstruct dense 3D shapes from sparse or partial point clouds. 
The network leverages point-wise feature aggregation and progressive refinement to capture both global shape priors 
and local geometric details, producing complete and smooth reconstructions. 

Our implementation of CRA-PCN is based on the official code provided in \url{https://github.com/EasyRy/CRA-PCN}

\subsubsection{Training Protocol}
CRA-PCN is trained for 420 epochs with an effective batch size of 8 and a gradient accumulation step of 1. 
The Adam optimizer is employed with an initial learning rate of $1\times10^{-3}$ and no weight decay. 
A GradualWarmup scheduler is adopted to stabilize training: it first performs gradual warm-up over the first 200 epochs and then applies a StepLR decay with a step size of 50 epochs 
and a decay factor of 0.5. The model is optimized using the Chamfer Distance as the training objective.

\subsection{ODGNet\cite{cai2024orthogonal}}

\subsubsection{Network Architecture}
ODGNet employs a Transformer-based hierarchical upsampling architecture to progressively reconstruct dense point clouds. 
The network generates 16,384 points from 1,024 input seeds using multi-scale upsampling factors of 
$[1, 4, 4]$. 
Each seed is encoded into a 128-dimensional feature vector, 
which is further processed within a 512-dimensional feature space. 
A learnable dictionary of 256 atoms facilitates feature expansion and refinement. 
The architecture balances both global and local geometric reasoning, enabling accurate shape reconstruction without L2 loss regularization.  
The loss function incorporates four equally weighted components: sparse loss, dense loss, $\mathrm{d}z$ loss, and orthogonality loss, 
each assigned a weight of 1.0.

Our implementation of ODGNet is based on the official code provided in \url{https://github.com/corecai163/ODGNet}

\subsubsection{Training Protocol}
ODGNet is trained for 420 epochs with an effective batch size of 8 and a gradient accumulation step of 1. 
The AdamW optimizer is adopted with an initial learning rate of $4\times10^{-4}$ and a weight decay of $5\times10^{-4}$. 
A LambdaLR scheduler decays the learning rate by a factor of 0.8 every 20 epochs, with a minimum learning rate set to 1\% of the initial value. 
Additionally, a batch normalization momentum scheduler (BNM scheduler) decreases the momentum by a factor of 0.5 every 20 epochs, 
starting from 0.9 and capped at a lower bound of 0.01. The model is optimized using the Chamfer Distance as the main evaluation metric.

\subsection{PointAttN \cite{wang2024pointattn}}

\subsubsection{Network Architecture}
PointAttN employs an attention-based encoder–decoder architecture tailored for point cloud completion. 
The network predicts dense point sets of 2,048 points from partial inputs by leveraging point-wise attention 
mechanisms to capture long-range dependencies and structural correlations within the 3D space. 
This design allows the model to preserve fine geometric details while maintaining global shape consistency 
through attentive feature fusion.

Our implementation of PointAttN is based on the official code provided in \url{https://github.com/ohhhyeahhh/PointAttN}

\subsubsection{Training Protocol}
PointAttN is trained for 420 epochs with a batch size of 8 using the Adam optimizer, where $\beta_1 = 0.9$ and $\beta_2 = 0.999$ and an initial learning rate of $1\times10^{-4}$ without weight decay. 
A learning rate decay strategy is employed, reducing the learning rate by a factor of 0.7 every 40 epochs, 
with a lower bound of $1\times10^{-6}$. 
Training and validation losses are monitored at each epoch, and the model checkpoint achieving the lowest 
validation Chamfer Distance is selected for final evaluation.

\subsection{SymmCompletion}

\subsubsection{Network Architecture}
SymmCompletion adopts a hierarchical upsampling architecture that progressively refines incomplete point clouds. The network incorporates two upsampling stages with factors of $2$ and $8$, enabling coarse-to-fine reconstruction of the underlying geometry. Notably, the input points are excluded during the reconstruction stage to emphasize feature-driven completion and promote effective learned feature propagation.

Our implementation of SymmCompletion is based on the official code provided in \url{https://github.com/HKUST-SAIL/SymmCompletion}

\subsubsection{Training Protocol}
SymmCompletion is trained for 420 epochs with an effective batch size of 8 and a gradient accumulation step of 1. 
The AdamW optimizer is employed with an initial learning rate of $2\times10^{-4}$ and a weight decay of $5\times10^{-4}$. 
A WarmUpCosLR scheduler linearly increases the learning rate during the first 20 warm-up epochs, 
followed by a cosine annealing schedule that decays the learning rate from $2\times10^{-4}$ to $1\times10^{-5}$ 
over the remaining epochs. The model is optimized using the Chamfer Distance as the loss function.

\subsection {PCCN~\cite{beetz2023multi}}
\subsubsection{Dataset}

PCCN is trained and evaluated on the same dataset introduced in Section~\ref{sec:dataset}. Each training sample contains a resampled sparse input consisting of 7,500 points and a structured dense ground truth composed of 6 classes, each resampled to 2,700 points.

\subsubsection{Network Architecture}

The PCCN model is capable of producing two outputs: a coarse output with shape $[B \times 7500 \times 4]$ and a dense output with shape $[B \times 2700 \times 6 \times 3]$. Both outputs are reshaped and concatenated before visualization or metric computation.

\subsubsection {Loss Function}
\label{sec:Loss Function}

The model is optimized using the Semantic-Aware Chamfer Distance (SA-CD) loss described in Sec.~\ref{sec:loss}. The total loss used during training is defined as:
\begin{equation}
\mathcal{L}_{\text{total}} = \mathcal{L}_{\text{coarse}} + \alpha \cdot \mathcal{L}_{\text{fine}},
\end{equation}
where $\mathcal{L}_{\text{coarse}}$ is the SA-CD between the coarse output and GT. $\mathcal{L}_{\text{fine}}$ is the SA-CD between the fine output and GT.

The weight $\alpha$ is gradually increased throughout training, following a step-wise schedule:

\begin{center}
\begin{tabular}{@{}lc@{}}
\toprule
Training Steps & $\alpha$ \\
\midrule
0 -- 10{,}000 & 0.01 \\
10{,}000 -- 20{,}000 & 0.1 \\
20{,}000 -- 40{,}000 & 0.5 \\
40{,}000 -- 80{,}000 & 1.0 \\
80{,}000 -- 160{,}000 & 5.0 \\
160{,}000 -- 320{,}000 & 10.0 \\
320{,}000 -- 640{,}000 & 50.0 \\
$>$640{,}000 & 100.0 \\
\bottomrule
\end{tabular}
\end{center}
This progressive weighting encourages the network to first learn global anatomical structures via coarse predictions, and later focus on finer structural details.

\subsubsection{Training Protocol}


The Adam optimizer is employed with a learning rate of $1 \times 10^{-4}$, where $\beta_1 = 0.9$ and $\beta_2 = 0.999$ control the exponential decay rates for the first and second moment estimates, respectively. These parameters ensure stable and adaptive gradient updates throughout training. StepLR is used, with a decay factor of $\gamma = 0.7$ applied every 50 epochs to gradually reduce the learning rate. The model is trained with a batch size of 8 for a total of 420 epochs.

The evaluation, and logging follow the same setup as described in Section~\ref{sec:Evaluation and Logging}.

\section{Additional Experimental Analyses}
\label{sec:Additional_Experimental_Analyses}

\subsection{Additional Evaluation on HeartCompv1 dataset}

\begin{figure*}[!t]
\centering
\includegraphics[width=0.95\linewidth]{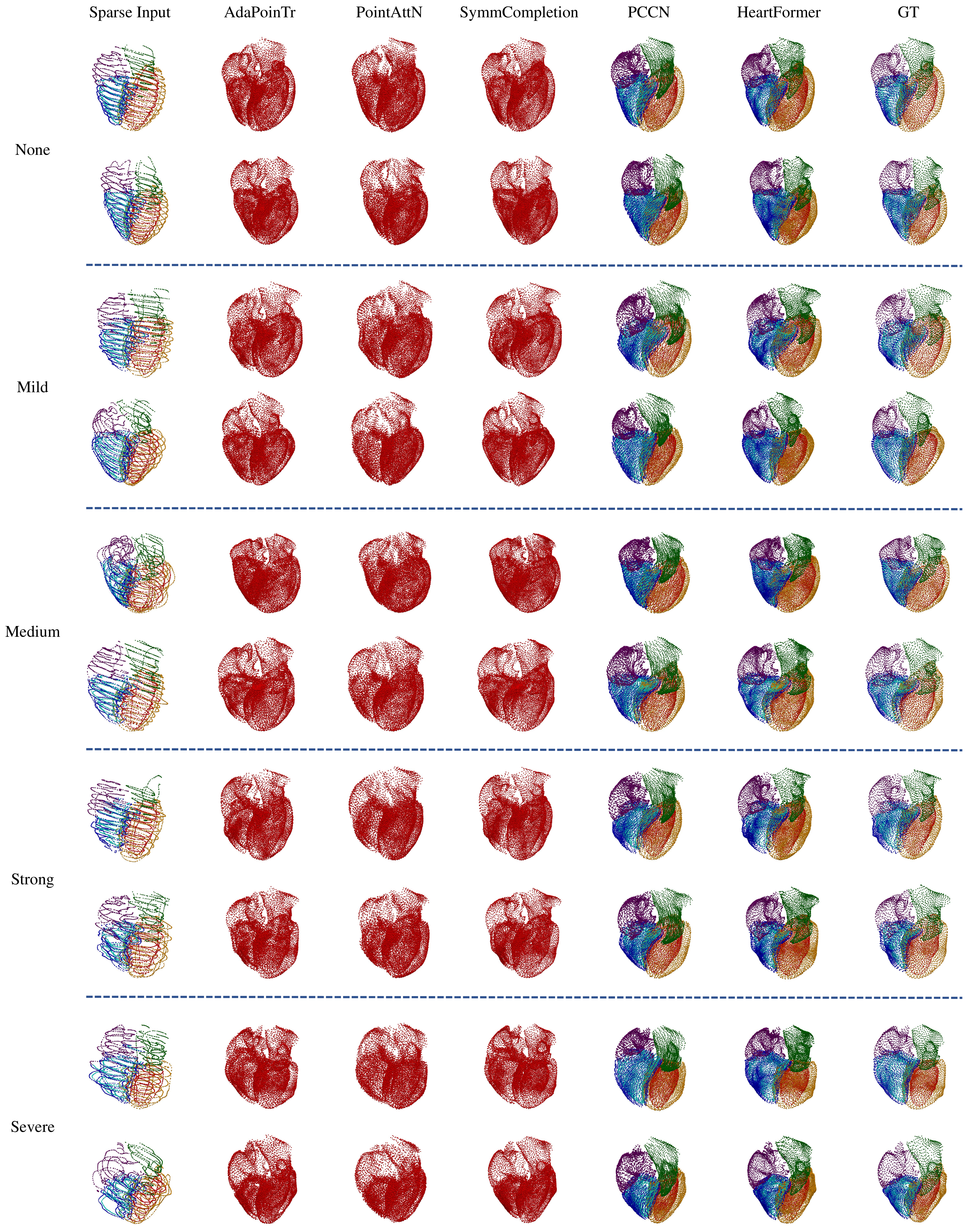} 
\caption{\textbf{Comparison of reconstruction results under five different levels of misalignment in the HeartCompv1 dataset.} Sparse Input denotes the partial point clouds with varying degrees of misalignment. AdaPoinTr, PointAttN, and SymmCompletion are the single-class point cloud completion models, while PCCN and the proposed HeartFormer are multi-class completion models. GT indicates the ground-truth.
}
\label{fig_11}
\end{figure*}

We present a quantitative comparison of reconstruction performance on SSD across five levels of misalignment in the HeartCompv1 dataset in Tab.~\ref{tab_11}, serving as supplementary results for the experiments in Sec.~\ref{sec:Evaluation_HeartCompv1}. Our HeartFormer consistently outperforms other state-of-the-art methods under all misalignment conditions. In addition, Fig.~\ref{fig_11} provides further qualitative comparisons across the same five misalignment levels. As shown, HeartFormer produces more detailed reconstructions and accurately preserves the semantic labels of cardiac substructures. The reconstructed details of our methods are clearly superior to those produced by multi-class completion methods, PCCN. We also provide a video in supplementary materials illustrating the completion training process, offering a visualized view of how the reconstruction progressively improves during training.

\begin{table}[]
\caption{\textbf{Quantitative comparison of reconstruction results for five different levels of misalignment in the HeartCompv1 dataset.}}
\centering
\resizebox{\linewidth}{!}{
\begin{tabular}{l|ccccc}
\midrule
\multirow{2}{*}{Methods} & \multicolumn{5}{c}{SSD$\downarrow$}                  \\
                         & None  & Mild  & Medium & Strong & Severe \\ \midrule
FoldingNet \cite{yang2017foldingnet}        & 2.041 & 2.130 & 2.272  & 2.412  & 2.591  \\
PCN \cite{yuan2018pcn}               & 1.936 & 2.003 & 2.170  & 2.391  & 2.684  \\
SnowflakeNet \cite{xiang2022snowflake}      & 2.194 & 2.400 & 2.913  & 3.538  & 4.287  \\
PoinTr \cite{yu2021pointr}            & 1.157 & 1.472 & 1.763  & 2.051  & 2.386  \\
FBNet \cite{yan2022fbnet}             & 1.602 & 1.847 & 2.076  & 2.298  & 2.543  \\
AdaPoinTr \cite{yu2301adapointr}         & 1.350 & 1.534 & 1.746  & 1.967  & 2.210  \\
AnchorFormer \cite{chen2023anchorformer}      & 1.543 & 1.991 & 2.455  & 2.976  & 3.614  \\
CRA-PCN \cite{rong2024cra}           & 1.630 & 1.938 & 2.281  & 2.759  & 3.410  \\
ODGNet \cite{cai2024orthogonal}            & 1.529 & 1.771 & 2.016  & 2.341  & 2.767  \\
PointAttN \cite{yu2021pointr}         & 1.727 & 1.871 & 2.060  & 2.269  & 2.510  \\
SymmCompletion \cite{yan2025symmcompletion}    & 1.320 & 1.470 & 1.664  & 1.875  & 2.086  \\
PCCN \cite{beetz2023multi}              & 1.634 & 1.739 & 1.944  & 2.201  & 2.497  \\
HeartFormer (Our)          & \textbf{1.289} & \textbf{1.454} & \textbf{1.655}  & \textbf{1.861}  & \textbf{2.072}  \\ \midrule
\end{tabular}
\label{tab_11}
}
\end{table}

\subsection{Additional Evaluation on UK Biobank dataset}

We provide a quantitative comparison against PCCN on the UK Biobank dataset using SSD across all six cardiac substructures in Tab.~\ref{tab_11}. Our model consistently and substantially outperforms PCCN on every substructure, highlighting its robustness and improved capability to capture fine-grained anatomical details.

\begin{figure*}[!t]
\centering
\includegraphics[width=\linewidth]{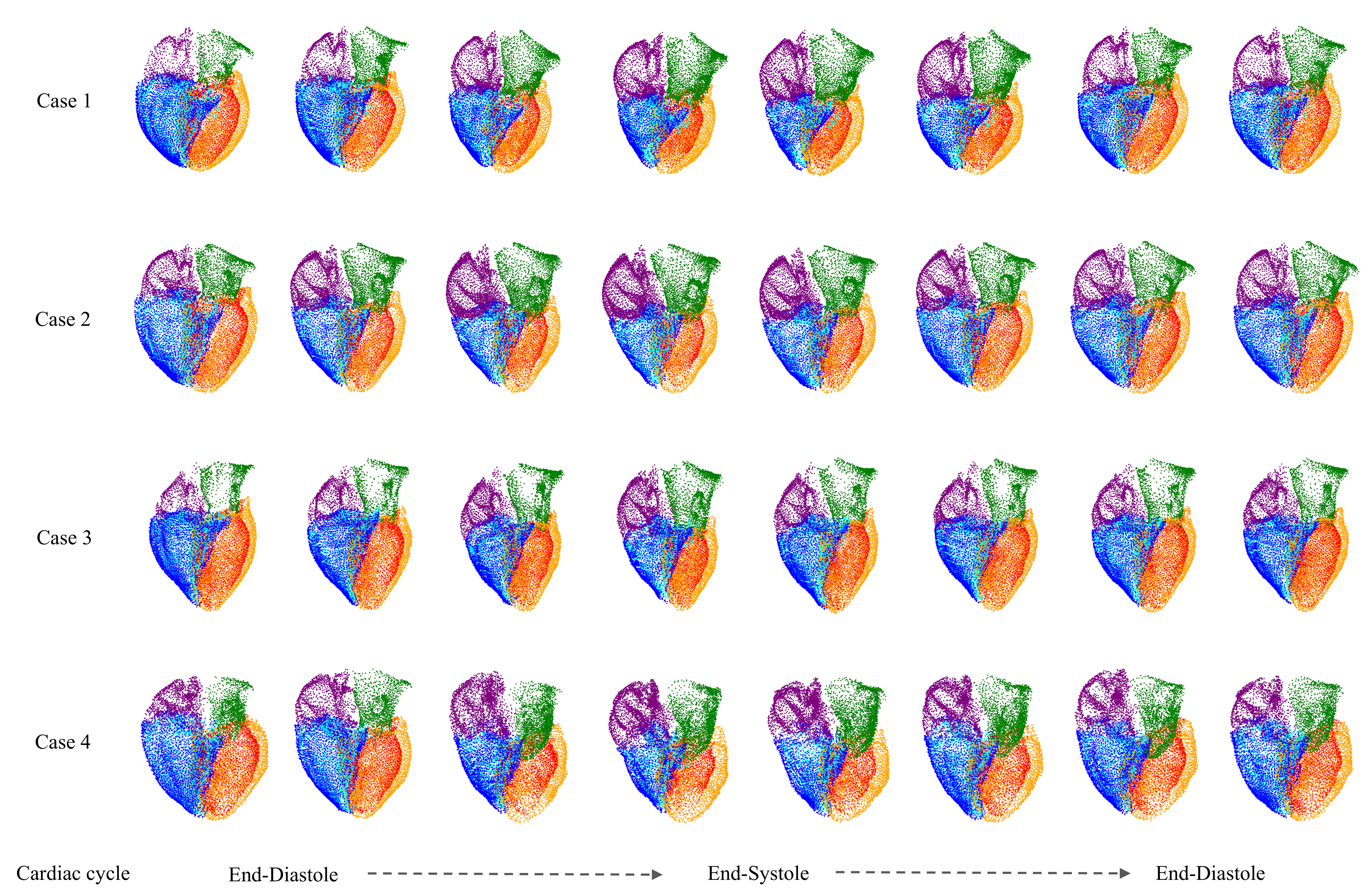} 
\caption{\textbf{Point cloud reconstruction results on the UK Biobank dataset.} 4D point cloud reconstructions for four representative subjects, demonstrating that the model captures clear systolic–diastolic dynamics, preserves smooth temporal deformation, and maintains anatomically coherent chamber evolution throughout the cardiac cycle.
}
\label{fig_12}
\end{figure*}

To further assess reconstruction fidelity, we present 4D point cloud reconstructions for four representative subjects. As illustrated in Fig.~\ref{fig_12}, our model preserves clear systolic and diastolic motion across all chambers, exhibiting smooth temporal dynamics and anatomically coherent shape evolution. We also provide videos for three representative cases in the supplementary materials. The videos can more provide an intuitive representation of cardiac motion and demonstrate that our method can reliably recover both global deformation patterns and subtle local structural variations. 

\begin{figure*}[!t]
\centering
\includegraphics[width=\linewidth]{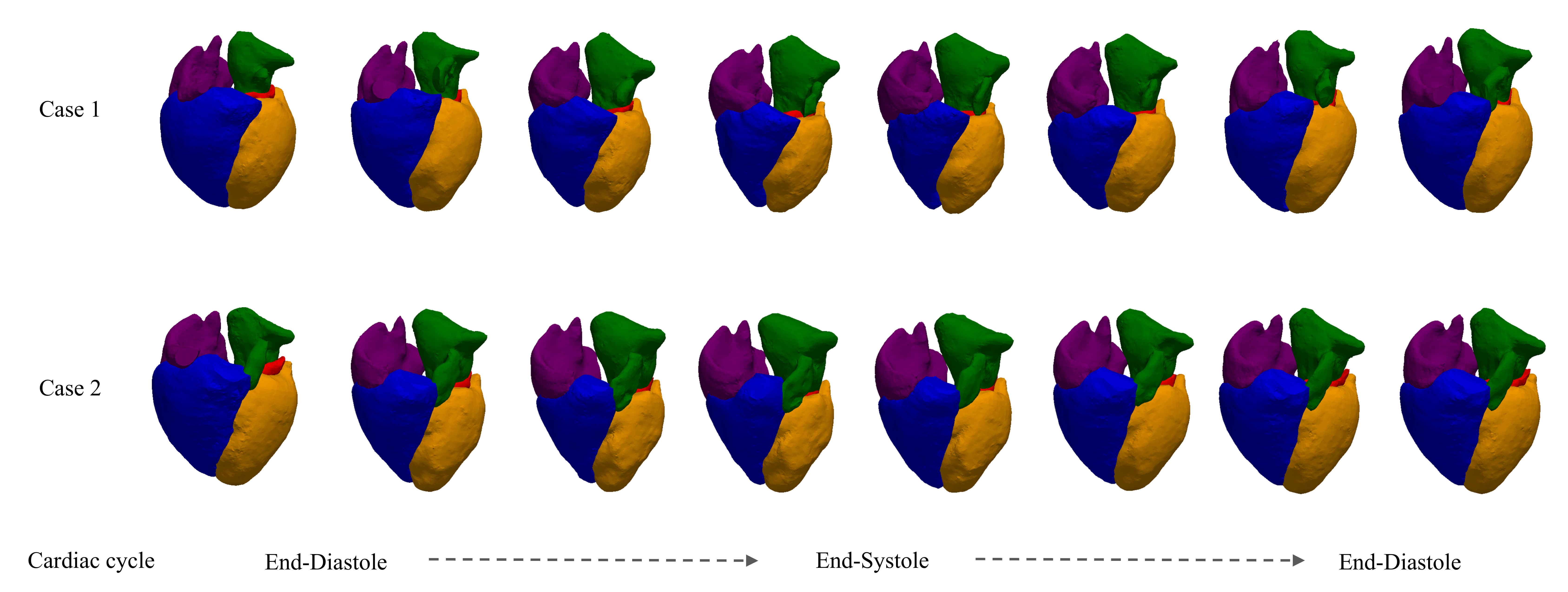} 
\caption{\textbf{Mesh reconstruction results on the UK Biobank dataset.}
4D mesh reconstructions for two representative cases, highlighting the temporal progression of cardiac morphology and enabling clear visualization of chamber contraction and relaxation. The results indicate that the model produces physiologically plausible cardiac motion while preserving detailed anatomical structure over time.}
\label{fig_13}
\end{figure*}

In addition, we provide mesh reconstruction results for two cases in Fig.~\ref{fig_13}. The mesh visualizations further highlight the temporal changes of cardiac morphology, allowing us to clearly observe the progression of cardiac motion throughout the cardiac cycle. Together, these qualitative results reinforce the capability of our framework to generate physiologically meaningful 4D cardiac reconstructions.

We additionally compute the ejection fraction (EF) for healthy subjects, obtaining an average value of 58.67\%. This result falls well within clinically normal ranges, suggesting that the reconstructed dynamics are physiologically plausible. Together, these quantitative and qualitative findings indicate that our model not only achieves superior reconstruction performance but also holds strong potential for downstream clinical analysis and real-world deployment.

\begin{table}[]
\caption{\textbf{Quantitative comparison of different models on the UK Biobank.} HeartFormer significantly outperforms PCCN across all six cardiac substructures under SSD metric.}
\centering
\resizebox{\linewidth}{!}{
\begin{tabular}{c|cccccc}
\midrule
\multirow{2}{*}{Methods} & \multicolumn{6}{c}{SSD$\downarrow$}                             \\
                         & LV endo & LV epi & RV endo & RV epi & LA    & RA    \\ \midrule
PCCN                     & 2.717   & 2.993  & 3.145   & 3.207  & 6.536 & 4.232 \\
HeartFormer              & \textbf{1.425}   & \textbf{1.391}  & \textbf{1.586}   & \textbf{1.694}  & \textbf{1.686} & \textbf{1.768} \\ \midrule
\end{tabular}
}
\label{tab_12}
\end{table}

\section{Additional Ablation Studies}
\label{sec:Additional_Ablation_Studies}

\begin{table}[]
\caption{\textbf{Quantitative comparison between our SA-DSTNet and adaption of previous initial point cloud generation networks on HeartCompv1 dataset.}}
\centering
\resizebox{0.9\linewidth}{!}{
\begin{tabular}{cc|ccc}
\midrule
\multicolumn{2}{c|}{Initial Method} & \multicolumn{3}{c}{Evaluation matrix} \\ 
SA-DSTNet        & SA-LSTNet        & CD $\downarrow$          & HD $\downarrow$         & SSD $\downarrow$        \\ \midrule
                 & \cmark                & 1.652       & 7.216      & 1.719      \\
\cmark                &                  & 1.596       & 6.750       & 1.661      \\ \midrule
\end{tabular}
}
\label{tab_13}
\end{table}

\begin{table}[]
\caption{\textbf{Quantitative evaluation of the Jitter Layer in GSBlock and the SA-Adaptive Sampling module in SSBlock on the HeartCompv1 dataset.}}
\centering
\resizebox{0.9\linewidth}{!}{
\begin{tabular}{cc|ccc}
\midrule
\multicolumn{1}{c|}{GSBlock}      & SSBlock              & \multicolumn{3}{c}{Evaluation matrix} \\
\multicolumn{1}{c|}{Jitter Layer} & SA-Adaptive Sampling & CD $\downarrow$          & HD $\downarrow$        & SSD $\downarrow$        \\ \midrule
\cmark                                 &                      & 1.645       & 7.073      & 1.701      \\
                                  & \cmark                    & 1.623       & 6.967      & 1.688      \\
\cmark                                 & \cmark                    & 1.596       & 6.750       & 1.661      \\ \midrule
\end{tabular}
}
\label{tab_15}
\vspace{-4mm}
\end{table}

\begin{table}[]
\vspace{-2mm}
\caption{\textbf{Quantitative comparison between our SA-GFRTNet and adaption of previous refinement module on HeartCompv1 dataset.}}
\centering
\resizebox{\linewidth}{!}{
\begin{tabular}{ccc|ccc}
\midrule
\multicolumn{3}{c|}{Refinement Module} & \multicolumn{3}{c}{Evaluation matrix} \\
SA-SDG   & SA-SGFormer   & SA-GFRTNet  & CD $\downarrow$          & HD $\downarrow$        & SSD $\downarrow$       \\ \midrule
\cmark         &               &             & 1.686       & 7.383      & 1.789      \\
         & \cmark              &             & 1.646       & 7.255      & 1.692      \\
         &               & \cmark            & 1.596       & 6.750       & 1.661      \\ \midrule
\end{tabular}
}
\label{tab_16}
\vspace{-4mm}
\end{table}

In this section, we provide extra ablation studies for our HeartFormer.

\subsection{Ablation for SA-DSTNet.}

\noindent\textbf{Ablation for components in GSBlock and SSBlock.}

Table~\ref{tab_15} reports the quantitative ablation results for the Jitter Layer in GSBlock and the SA-Adaptive Sampling module in SSBlock on the HeartCompv1 dataset. Introducing either component individually yields consistent performance gains across all three metrics, demonstrating that both modules effectively enhance geometric fidelity. Notably, enabling SA-Adaptive Sampling leads to clear improvements over the baseline without it, confirming its ability to better capture local structural variations. When both the Jitter Layer and SA-Adaptive Sampling are used jointly, the model achieves the lowest errors in all metrics, indicating that the two components are complementary and together deliver the most accurate reconstruction.

\noindent\textbf{Ablation for initial point cloud generation networks.}
In Tab~\ref{tab_13}, we present a quantitative comparison between our SA-DSTNet and adapted variants of prior initial point cloud generation networks on the HeartCompv1 dataset. We extend DSTNet with semantic-aware capability to better handle anatomical substructures. As shown in the Tab~\ref{tab_13}, our SA-DSTNet consistently outperforms adaption LSTNet across all evaluation metrics.

\subsection{Ablation for SA-GFRTNet.}

Table~\ref{tab_16} presents a quantitative comparison of our SA-GFRTNet against semantically adapted variants of prior refinement modules on the HeartCompv1 dataset. Incorporating semantic awareness into either the SDG module in SVD-Former~\cite{zhu2023svdformer} or the SGFormer module in SymmCompletion~\cite{yan2025symmcompletion} enables these methods to perform semantically guided refinement and achieve multi-class completion of cardiac substructures. However, our SA-GFRTNet achieves the best performance across all metrics, outperforming both SA-SDG and SA-SGFormer by a clear margin. These results demonstrate that the proposed geometric–feature refinement mechanism in SA-GFRTNet provides a more effective refinement strategy, enabling more accurate and semantically consistent reconstruction of cardiac substructures.

\subsection{Ablation for Loss Function.}
\begin{figure}[!t]
\centering
\resizebox{\linewidth}{!}{
\includegraphics[width=3.5in]{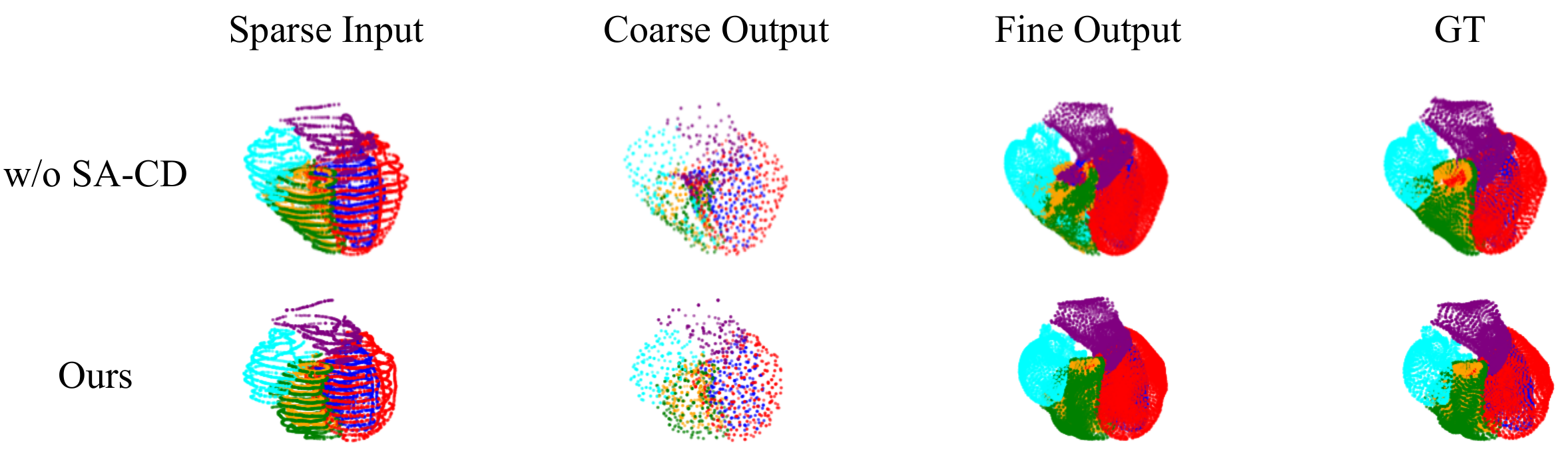} 
}
\caption{Comparison of reconstruction results with and without the Semantic-Aware Geometry loss. Incorrect semantic labeling can be clearly observed in the model trained without Semantic-Aware guidance.}
\label{fig_loss}
\vspace{-4mm}
\end{figure}

In Fig.~\ref{fig_loss}, we present a comparison of reconstruction results with and without the Semantic-Aware Geometry loss. The model trained without semantic-aware guidance exhibits evident semantic inconsistencies, including incorrect anatomical labeling and structurally implausible boundaries. In contrast, incorporating the Semantic-Aware Geometry loss leads to substantially cleaner anatomical separation and more faithful geometric reconstruction. We also provide the quantitative results in Tab.~\ref{tab7} of the main paper. These results demonstrate that semantic-aware supervision is essential for achieving anatomically consistent and reliable shape reconstruction.

\section{Code Implementation}
\label{sec:Code}

The source code and pre-trained models are included in the supplementary materials. A detailed README.md file is provided to facilitate reproduction.
Due to the size constraint of the supplementary materials, the complete dataset (12.5 GB) cannot be included. The full dataset, together with the code, will be released on our GitHub repository upon acceptance.

To support local data generation, we provide the pseudocode of the synthetic dataset generation pipeline in Section~\ref{sec:SSM}.
For convenience, we also include 10 representative test samples in the supplementary for evaluation.

\section{Discussion}
\label{sec:Discussion}


Our current framework can generate 4D reconstructions of the four-chamber heart and produce time-resolved videos that capture the dynamic motion of each atrial and ventricular structure. We further compute clinically relevant metrics, including chamber volume and ejection fraction, to validate the physiological plausibility of the reconstructed cardiac motion, demonstrating the potential of our method for downstream clinical assessment. 

To the best of our knowledge, existing cardiac reconstruction pipelines invariably rely on segmentation masks produced by separate models for 3D cardiac geometry reconstruction. Although HeartFormer demonstrates strong performance, it currently depends on this external step. In future work, we plan to integrate HeartFormer with a segmentation backbone to enable end-to-end reconstruction of four-chamber cardiac geometry directly from 2D cine-MRI, further streamlining the pipeline and enhancing clinical applicability.

\end{document}